\documentclass{article}
\usepackage[justification=centering]{caption}
\usepackage[numbers]{natbib}
\usepackage{PRIMEarxiv}
\usepackage[utf8]{inputenc} 
\usepackage[T1]{fontenc}    
\usepackage{hyperref}       
\usepackage{url}            
\usepackage{booktabs}       
\usepackage{amsfonts}       
\usepackage{nicefrac}       
\usepackage{microtype}      
\usepackage{lipsum}
\usepackage{fancyhdr}       
\usepackage{graphicx}       
\graphicspath{{media/}}     

\title{A systematic review of Hate Speech automatic detection using Natural Language Processing}                      

\author{
  Md Saroar Jahan, and Mourad Oussalah \\
  University of Oulu, CMVS, BP 4500, 90014 Finland \\
  \texttt{\{mjahan18,mourad.oussalah\}@oulu.fi} \\
}

\begin{document}
\maketitle

\begin{abstract}
With the multiplication of social media platforms, which offer anonymity, easy access and online community formation and online debate, the issue of hate speech detection and tracking becomes a growing challenge to society, individual, policy-makers and researchers. Despite efforts for leveraging automatic techniques for automatic detection and monitoring, their performances are still far from satisfactory, which constantly calls for future research on the issue. This paper provides a systematic review of literature in this field, with a focus on natural language processing and deep learning technologies, highlighting the terminology, processing pipeline, core methods employed, with a focal point on deep learning architecture. From a methodological perspective, we adopt PRISMA guideline of systematic review of the last 10 years literature from  ACM Digital Library and  Google Scholar. In the sequel, existing surveys, limitations, and future research directions are extensively discussed.

\end{abstract}



\maketitle

\section{Introduction}
In the era of social computing, the interaction between individuals becomes more striking, especially through social media platforms and chat forums. Microblogging applications opened up the chance for people worldwide to express and share their thoughts instantaneously and extensively. Driven, on one hand, by the platform's easy access and anonymity. And, on the other hand, by the user's desire to dominate debate, spread / defend opinions or argumentation, and possibly some business incentives, this offered a fertile environment to disseminate aggressive and harmful content. Despite the discrepancy in hate speech legislation from one country to another, it is usually thought to include communications of animosity or disparagement of an individual or a group on account of a group characteristic such as race, color, national origin, sex, disability, religion, or sexual orientation  \cite{nockleby2000hate}.
Benefiting from the variation in national hate speech legislation, the difficulty to set a limit to the constantly evolving cyberspace, the increased need of individuals and societal actors to express their opinions and counter-attacks from opponents and the delay in manual check by internet operators, the propagation of hate speech online has gained new momentum that continuously challenges both policy-makers and research community.          
With the development in natural language processing (NLP) technology, much research has been done concerning automatic textual hate speech detection in recent years. A couple of renowned competitions (e.g.,  SemEval-2019\cite{zampieri2019semeval} and 2020 \cite{zampieri2020semeval}, GermEval-2018 \cite{wiegand2018overview}) have held various events to find a better solution for automated hate speech detection. In this regard, researchers have populated large-scale datasets from multiple sources, which fueled research in the field. Many of these studies have also tackled hate speech in several non-English languages and online communities. This led to investigate and contrast various processing pipelines, including the choice of feature set and Machine Learning (ML) methods (e.g., supervised, unsupervised, and semi-supervised), classification algorithms (e.g., Naives Bayes, Linear Regression, Convolution Neural Network (CNN), LSTM, BERT deep learning architectures, and so on). The limitation of the automatic textual-based approach for efficient detection has been widely acknowledged, which calls for future research in this field. Besides, the variety of technology, application domain, and contextual factors require a constant up-to-date of the advance in this field in order to provide the researcher with a comprehensive and global view in the area of automatic HT detection. Extending existing survey papers in this field, this paper contributes to this goal by providing an updated systematic review of literature of automatic textual hate speech detection with a special focus on machine learning and deep learning technologies. 
We frame the problem, its definition and identify methods and resources employed in HT detection. We adopted a systematic approach that critically analyses theoretical aspects and practical resources, such as datasets, methods, existing projects following PRISMA guidelines \cite{moher2009preferred}. In this regards, we have tried to answer the following research questions:

\begin{itemize}
    \item Q1: What are the specificities among different HS branches and scopes for automatic HS detection from previous literature?

 \item Q2: What is the state of the deep learning technology in automatic HS detection in practice?

    \item Q3: What is the state of the HS datasets in practice?
\end{itemize}

The above-researched questions will examine barriers and scopes for the automatic hate speech detection technology. A systematic review-based approach is conducted to answer Q1 and Q2, where we will try to depict and categorize the existing technology and literature. The third research question Q3, will be answered by critically examining the scope and boundaries of the dataset identified by our literature review, highlighting the characteristics and aspects of the available resources.

This review paper is organized as follows: section~\ref{background} will include a brief theoretical definition of HS. Section~\ref{Related_work} examines the previously identified review papers of HS detection. Section~\ref{METHODOLOGY} details the systematic literature review document collection methodology. Section  \ref{results} presents the results of this literature review, including the state of deep learning technology. Section~\ref{resources} emphasizes on the available resources (datasets and open-source projects). After that, in section  \ref{challenges}, an extensive discussion is carried out. Finally, we have highlighted future research directions and conclusions at the end of this paper.

\section{Background}
\label{background}
\subsection{What is hate speech?}
Deciding if a portion of text contains hate speech is not simple, even for human beings. Hate speech is a complex phenomenon, intrinsically associated with relationships between groups, and relies on language nuances. Different organization and authors have tried to define hate speech as follow:

\begin{enumerate}
    \item \textbf{Code of Conduct between  European Union Commission and companies: } "All conduct publicly inciting to violence or hatred directed against a group of persons or a member of such a group defined by reference to race, color, religion, descent or national or ethnic" \cite{wigand2017speech}.
  
    \item \textbf{International minorities associations (ILGA) :} 'Hate crime is any form of crime targeting people because of their actual or perceived belonging to a particular group. The crimes can manifest in a variety of forms: physical and psychological intimidation, blackmail, property damage, aggression and violence, rape'\footnote{\url{https://www.ilga-europe.org/what-we-do/our-advocacy-work/hate-crime-hate-speech}}.
    
    \item \textbf{\citeauthor{nobata2016abusive} } \cite{nobata2016abusive}- "Language which attacks or demeans a group based on race, ethnic origin, religion, disability, gender, age, disability, or sexual orientation/gender identity."
    
    \item \textbf{Facebook:} "We define hate speech as a direct attack against people on the basis of what we call protected characteristics: race, ethnicity, national origin, disability, religious affiliation, caste, sexual orientation, sex, gender identity, and serious disease. We define attacks as violent or dehumanizing speech, harmful stereotypes, statements of inferiority, expressions of contempt, disgust or dismissal, cursing, and calls for exclusion or segregation. We consider age a protected characteristic when referenced along with another protected characteristic. We also protect refugees, migrants, immigrants, and asylum seekers from the most severe attacks, though we do allow commentary and criticism of immigration policies. Similarly, we provide some protections for characteristics like occupation, when they're referenced along with a protected characteristic\footnote{\url{https://www.facebook.com/communitystandards/hate_speech}}."
    
    \item \textbf{Twitter: } 'You may not promote violence against, threaten, or harass other people on the basis of race, ethnicity, national origin, caste, sexual orientation, gender, gender identity, religious affiliation, age, disability, or serious disease' \footnote{\url{https://help.twitter.com/en/rules-and-policies/twitter-rules\#hateful-conduct}}. Examples from Twitter hate-speech are: 
    
    \begin{itemize}
        \item "I'm glad this [violent event] happened. They got what they deserved [referring to persons with the attributes noted above]."
        \item "[Person with attributes noted above] are dogs" or "[person with attributes noted above] are like animals."
    \end{itemize}

    \item \textbf{YouTube:} 'We remove content promoting violence or hatred against individuals or groups based on any of the following attributes: age, caste, disability, ethnicity, gender identity and expression, nationality, race, immigration status, religion, sex/gender, sexual orientation, victims of a major violent event and their kin, and veteran Status'. \footnote{\url{https://support.google.com/youtube/answer/2801939?hl$=$en}}
\end{enumerate}

To better understand the definitions, we consider different terms analysis from the above definitions.  Tab.~\ref{Tab:comparison_hatespeech_word} summarizes key components that characterize HS in these definitions. For instance, all the definitions point out that  HS  has specific targets (person,  group, nationality, etc.). Furthermore, most of these definitions refer to religion, gender discrimination, race, color, ethnicity, and violence.  While less common criteria for measuring HS include curse, disability, property damage, age, and serious disease.

\begin{table}
  \caption{Content Analysis of Hate Speech Definitions}
  \label{Tab:Content_Analysis_Speech}
  \centering
  \small
\begin{tabular}{|p{3cm}|p{1cm}|p{1.2cm}|p{1.3cm}|p{1.7cm}|p{2cm}|p{3cm}|}
\hline
\bf{Source} & \bf{Target} & \bf{Religion}  &\bf{Gender  Discrimination} & \bf{Race/ Ethnicity/ Color} & \bf{Radicalization/ Violence } & \bf{Others} \\ \hline 

EU Code of conduct & yes & yes & yes & yes & yes & - \\ \hline

ILGA  & yes & no & no & no & yes & intimidation, blackmail, property damage, rape \\ \hline

Scientific paper \cite{nobata2016abusive} & yes & yes & yes & yes & no & disability, age \\ \hline

Facebook & yes & yes & yes & yes & yes & curse, serious disease, Occupation \\ \hline

Twitter & yes & yes & yes & yes & no & age , disability, serious disease \\ \hline

YouTube & yes & yes & yes & yes & yes & age \\ \hline

\end{tabular}
\vspace{8mm}
\end{table}

\subsection{Other Related Concepts }
From the above definitions and contents analysis, it is clear that some elements are highly related to hate speech ( e.g., racism, violence, gender discrimination, etc.). Moreover, we have found several previous works that have presented significant branches of HS. The analysis of HS's different branches helps to reach insights from different perspectives. This expects to contribute to spotting and recognizing the interrelationships among these terminologies. Here we will discuss some of the essential categories that are found relevant in most Hate Speech studies:

\textbf{Cyberbullying:} \citeauthor{chen2012detecting} \cite{chen2012detecting}, \citeauthor{dinakar2012common} \cite{dinakar2012common} defined  the electronic form of traditional bullying, also, referred to cyberbullying, as the aggression and harassment targeted to an individual who is unable to defend himself. \citeauthor{dredge2014cyberbullying} opined that bullying is known for its repetitive act to the same individual, unlike hate speech which is more general and not necessarily intended to hurt a specific individual \cite{dredge2014cyberbullying}.

\textbf{Racism:}  This category includes racial offense or tribalism, regionalism, xenophobia (especially for migrant workers) and nativism (hostility against immigrants and refugees), and any prejudice against a particular tribe, region, color, or physical posture of an individual. For instance, offending an individual because he belongs to a specific tribe, region, or country \cite{Al-Hassan}

\textbf{Sexism, Gender discrimination:} \citeauthor{o2009encyclopedia}  \cite{o2009encyclopedia} defined sexism as a prejudice or a discrimination based on a person's sex or gender. Sexism can affect anyone, although, it primarily affects women and girls. It has also been linked to stereotypes and gender roles. \citeauthor{matsumoto2001handbook} \cite{matsumoto2001handbook} stated that in many types of hate, there could be the existence of sexual harassment contents. Moreover, \citeauthor{jha2017does} \cite{jha2017does} reported that sexism might come in two different forms: Hostile (which is an explicit negative attitude) and Benevolent (which is more subtle).

\textbf{Radicalization:} This concept is usually referred to as a motive towards violent extremism. Radicalization and hate speeches are closely related and sometimes used equivalently. Some authors link radicalization to religious-based hate speech. \citeauthor{wadhwa2013tracking} \cite{wadhwa2013tracking} referred to radical groups as "cyber-extremists." 

\textbf{Abusive language, Offensive language:}  The term abusive language refers to hurtful language and includes hate speech, derogatory language, and profanity. However, many researchers referred to the abusive language as offensive language \cite{nobata2016abusive}.

\textbf{Religious hate speech:}  This includes any religious discrimination, such as Islamic sects, calling for atheism, Anti-Christian, and their respective denominations or anti-Hinduism groups. \cite{Albadi2018are} reported that religious hate speech is considered a motive for crimes in countries with the highest social crimes.

\begin{figure}
    \centering
    \includegraphics[width=.8\linewidth,]{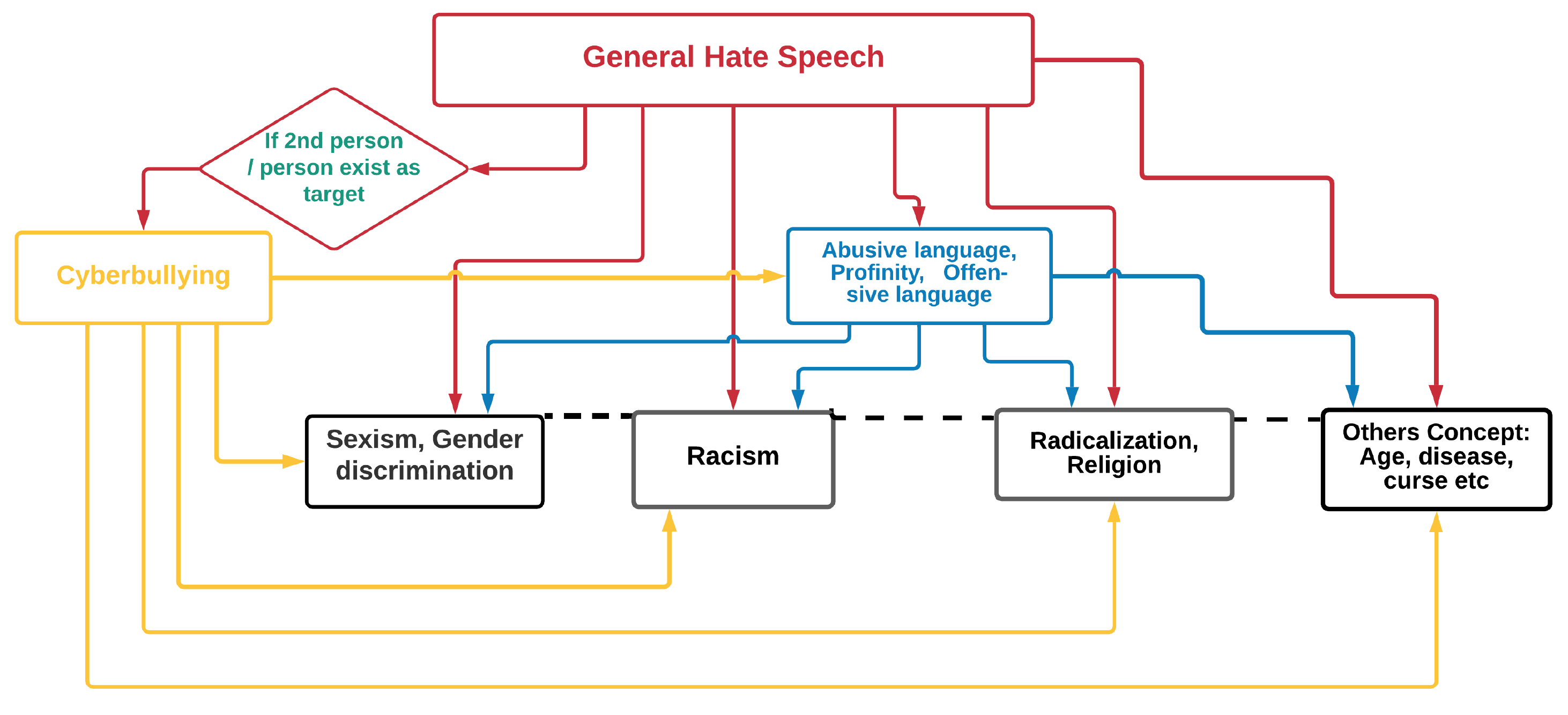}
    \caption{Relational diagram between different type of hate speech concepts.}
    \label{fig:HS_relation}
\end{figure}

\subsection{Relationship of HS Concepts and Example}
From the above definitions of general HS and other related concepts, we have drawn a relationship diagram shown in Fig.~\ref{fig:HS_relation}. 

All other related concepts, cyberbullying, racism, sexism, abusive language, and radicalization, have been derived from the HS concept, which acts as a parent node in this hierarchical construction. In the second level, cyberbullying and abusive/offensive concepts are distinguished. Next, other components like sexism, racism, radicalization are discerned. 

A critical claim advocated by some social science and psychiatry scholars stipulates that cyberbullying cases must include both Insult/Swear wording and a second person/person's name \cite{patchin2006bullies}. If there is no second person/person's name available, it may not be considered cyberbullying other than general HS.  This means that all cyberbullying cases can also be cast into the HS category, while some HS may not be classified as cyberbullying.

However, the above categorization is not always faithful, as exemplified in Table ~\ref{Tab:comparison_hatespeech_word}. For example, sentence-7 ("This is bad, but John Doe is lucky") includes both Insult word and Person name; however,  it is not an HS, cyberbullying, nor abusive case as the relationship between the two is not established.
Similarly, sentence-5, "John Doe is not bad," contains both Person name and Insult word with a clear link between the two entities, but it is not connected to HS, cyberbullying, nor abusive case due to the presence of negating form. Likewise,  sentence-4, 'John Doe is not a good person,' does not contain Insult words, but it is considered HS, cyberbullying, and offensive. In sentence-3, 'Asylum seekers are dirty,' has an Insult word and target; it falls into HS but is not categorized as cyberbullying because its target is not a Person / second person.

All these examples show the requirements mentioned above for HS, cyberbullying, and other cases are potential conditions for the occurrence of HS but not sufficient due to the complexity of the natural language modifiers expression that could negate or shift the meaning. 

Besides, recognition of HS may be boosted when multiple sentences were put together as in sentence-8 ("John doe working hard. Ugly"), which is considered as an HS, Cyberbullying and Offensive cases even though the second sentence "Ugly" contains only an Insult word without any second person/Person entity.  

There could be examples of sentences that may comprise multiple HS concepts at the same time.  Sentence-1 shows that it contains all categories,  HS, cyberbullying,  abusive, gender discrimination, racism, and radicalization.

The above few cases explain the complications of detecting HS cases and determining their exact categorization, which involves examining all the paragraph's textual information.

\begin{table}
  \caption{Sentence classification into different HS categories.}
  \label{Tab:comparison_hatespeech_word}
  \centering
  \small
\begin{tabular}{|p{4cm}|p{1cm}|p{1.2cm}|p{1.3cm}|p{1.7cm}|p{.9cm}|p{1cm}|p{1cm}|}\hline
\bf{Sentence} &  \bf{General HS} & \bf{Cyber- bulling} & \bf{Abusive Offensive} &\bf{Sexisim/ Gender Dicsrimination} & \bf{Racism} & \bf{Radica-lization} & \bf{Insult/ Swear word}   \\ \hline 

1. John doe is <religion name>  from <nationality> dirty is a bisexual and violent & yes & yes & yes & yes & yes & yes & exist \\ \hline

2. John doe is  dirty  & yes & yes & yes & no & no & no & exist \\ \hline
3. Asylum seekers are  not good  & yes & no & no & no & no & no & not-exist\\ \hline

4. John Doe is not good person & yes & yes & yes & no & no & no & not-exist \\ \hline

5. John Doe is not bad & no & no &  no & no & no & no & exist\\ \hline

6. Group for blacks only! & no & no & no & no & no & no & exist \\ \hline

7. This is bad, but John Doe is lucky & no & no & no & no & no & no & exist\\ \hline
8. John doe working hard. Ugly & yes & yes & yes & no & no & no & exist\\ \hline

\end{tabular}
\vspace{8mm}
\end{table}

\begin{figure}
    \centering
    \includegraphics[width=1\linewidth,]{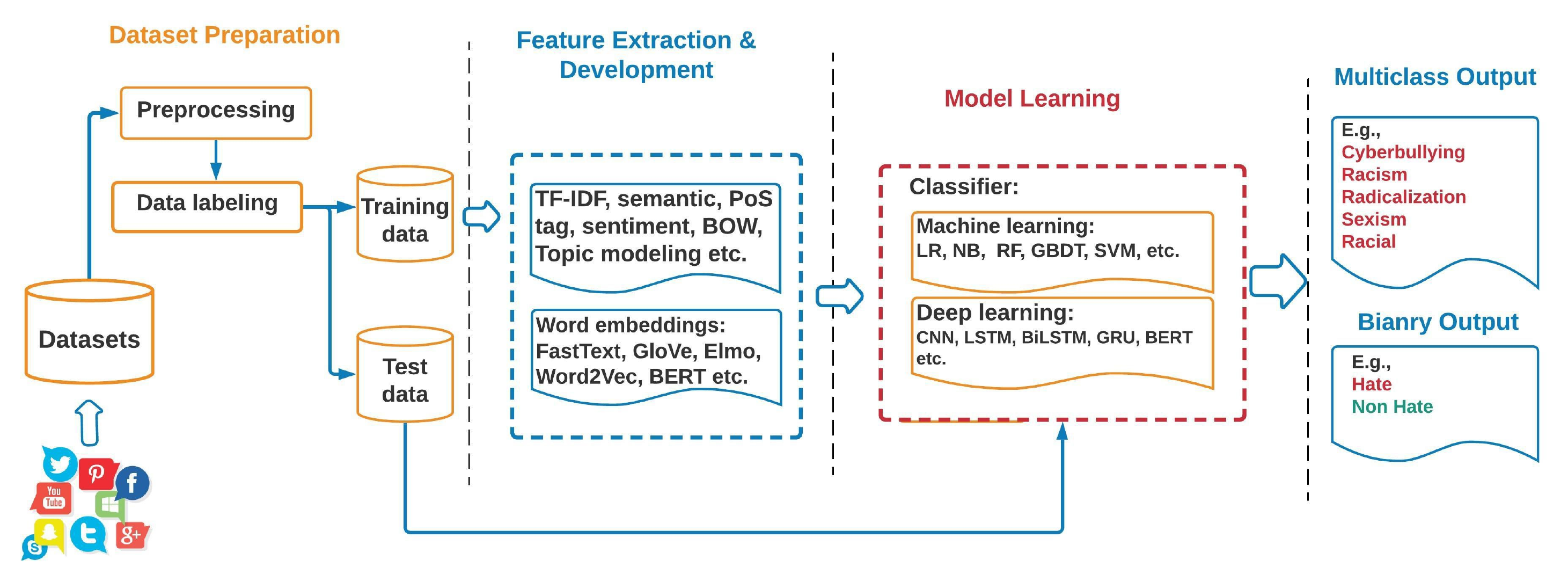}
    \caption{Generic pipeline of automatic HS detection system.}
    \label{fig:gp}
\end{figure}

\subsection{Generic Pipeline of Automatic HS Detection}

\citeauthor{kowsari2019text} \cite{kowsari2019text} stipulated four essential parts for any text classification task, which still are valid for the HS classification as well.  Figure \ref{fig:gp} highlights the generic pipeline of the HS detection task as a text classification system-based approach. Its main components are described below: 

\textbf{(i) Dataset collection and preparation:} is the first step is HS detection pipeline. Often, datasets are collected from social media platforms (Facebook, Youtube, Twitter, etc.). Preprocessing is performed according to dataset structure and quality. Typically, this involves filtering and normalization aspect of tuextual inputs, which include tokenization, stopwords removal, misspelling correction, noise removal, stemming, lemmatization, among others. 
We shall also notice that the dataset maybe provided initially so that no collection is required.  
As part of data preparation, training and testing parts of the dataset should be distinguished for the subsequent machine learning step.

\textbf{(ii) Feature Engineering:} is the next phase of the analysis where appropriate features are extracted from the textual inputs so that unstructured text sequences are converted into structured features. Common techniques for feature extractions are TF-IDF, semantic, lexical, topic modeling, sentiment, BOW, word embedding (FastText, GloVe, Word2Vec).  

Sometimes, dimensionality reduction is applied to reduce the time and memory complexity. Examples of dimension reduction methods are principal component analysis (PCA), linear discriminant analysis (LDA), non-negative matrix factorization (NMF), random projection, autoencoders, and t-distributed stochastic neighbor embedding (t-SNE) \cite{kowsari2019text}. 

\textbf{(iii) Model Training:} is one of the most crucial step of the text classification pipeline where a machine learning /deep learning model is trained on the training dataset. Several classifiers can be tailored based on task requirements: RF, NB, LR, CNN, RNN, BERT, etc. Commonly word embedding can be jointly used in a neural network model as an embedding layer which helps to enhance deep learning performance. The output of the machine-learning / deep-learning model can be either binary decision (e.g., hate versus non-hate speech) or multi-class output where the model discriminates various type of hate speech and non-hate speech.

\textbf{(iv) Evaluation:}  is the final part of the text classification pipeline where the performance of the machine learning /deep learning model is estimated. Several evaluation metrics are used for this purpose: accuracy, F1 score, precision, Matthews Correlation Coefficient (MCC), receiver operating characteristics (ROC), area under the ROC curve (AUC).

\begin{table}
  \caption{Acronyms and their full forms.}
  \label{Tab:abbreviation}
  \centering
  \small
\begin{tabular}{|p{1.5cm}|p{5cm}|p{1.5cm}|p{5cm}|}
\hline
\bf{Acronym} &  \bf{Full form} & \bf{Acronym} & \bf{Full form} \\ \hline 

CNN & Convolutional neural network & LSTM & Long short-term memory\\ \hline
GRU & Gated recurrent units. Its a gating mechanism in RNNN & BiLSTM & Bidirectional Long Short-Term Memory \\ \hline
BGRU &  Bidirectional gated recurrent unit network  & RNN & Recurrent neural \\\hline

BERT & Bidirectional Encoder Representations from Transformers & RoBERTa & A Robustly Optimized BERT Pretraining Approach\\ \hline
DistilBERT &  A distilled version of BERT: smaller, faster & ALBERT & A Lite BERT for Self-supervised Learning \\ \hline

mBERT & Multilingual BERT & TWilBert & A specialization of BERT architecture both for the Spanish  and the Twitter domain\\ \hline

LR & Linear regression & NB & Naive Bayes\\ \hline
RF & Random forest & SVM & Support Vector Machin \\\hline
TF-IDF & Term frequency-inverse document frequency & SG & Skip gram \\ \hline
GBDT & Gradient Boosting Decision Tree & ELMO & Embeddings from Language Models \\ \hline

BOW & Bag of word & CBOW & Continiuos bag of word \\\hline

PoS Tag & Part-of-speech tagging & GHSOM  & The Growing Hierarchical Self-Organizing Map \\\hline

\end{tabular}
\vspace{8mm}
\end{table}

\section{List of Acronyms}
In order to ease the readability and maintain the coherence of the various notations employed, we list in table~\ref{Tab:abbreviation} the various acronyms and their complete form cited throughout this paper. This concerns mainly the machine learning, deep learning, and feature sets of techniques reviewed in this paper.

\clearpage

\section{Related work}
\label{Related_work}
From a computer science point of view, the scientific study of hate speech is comparatively a new topic, for which the number of review papers in the field is limited. We found only a few survey/review articles during the process of literature review. These were obtained using a systematic review-based approach where we adopted PRISMA framework \cite{moher2009preferred} and conducted a brief systematic review of previous reviews in HS as will be detailed in the following section. 

\subsection{Methodology for collecting related  review papers}
\subsubsection{Keywords}
To collect related review papers, we first selected the best keywords to retrieve relevant information from the search database (Google scholar\footnote{\url{https://scholar.google.com/}} and ACM\footnote{\url{https://dl.acm.org/}}). Hate speech is a new concept that became popular recently; therefore, we considered other relevant terms referring to particular hate speech types (e.g., cyberbullying, sexism, racism, and homophobia). The search keywords were: Review/survey hate speech detection, Review/survey Offensive Or abusive language detection, Review/survey sexism detection, Review/survey sexism detection, and review/survey cyberbullying detection.

\subsubsection{Search for documents}
We utilized Google Scholar and ACM digital library in order to search for systematic reviews containing the term "review/survey' with the keywords mentioned above in their titles and abstracts. At the same time, no date and language restrictions were imposed. The reason for not selecting "systematic" as a search keyword is that we wanted to collect all reviews that were not only based on systematic methodology but also on narrative methods. The last search was run on 30  December 2020. The title, abstract, authors' names and affiliations, journal name, and year of publication of the identified records were exported to an MS Excel spreadsheet for further analysis.

\begin{figure}
    \centering
    \includegraphics[width=.8\linewidth,]{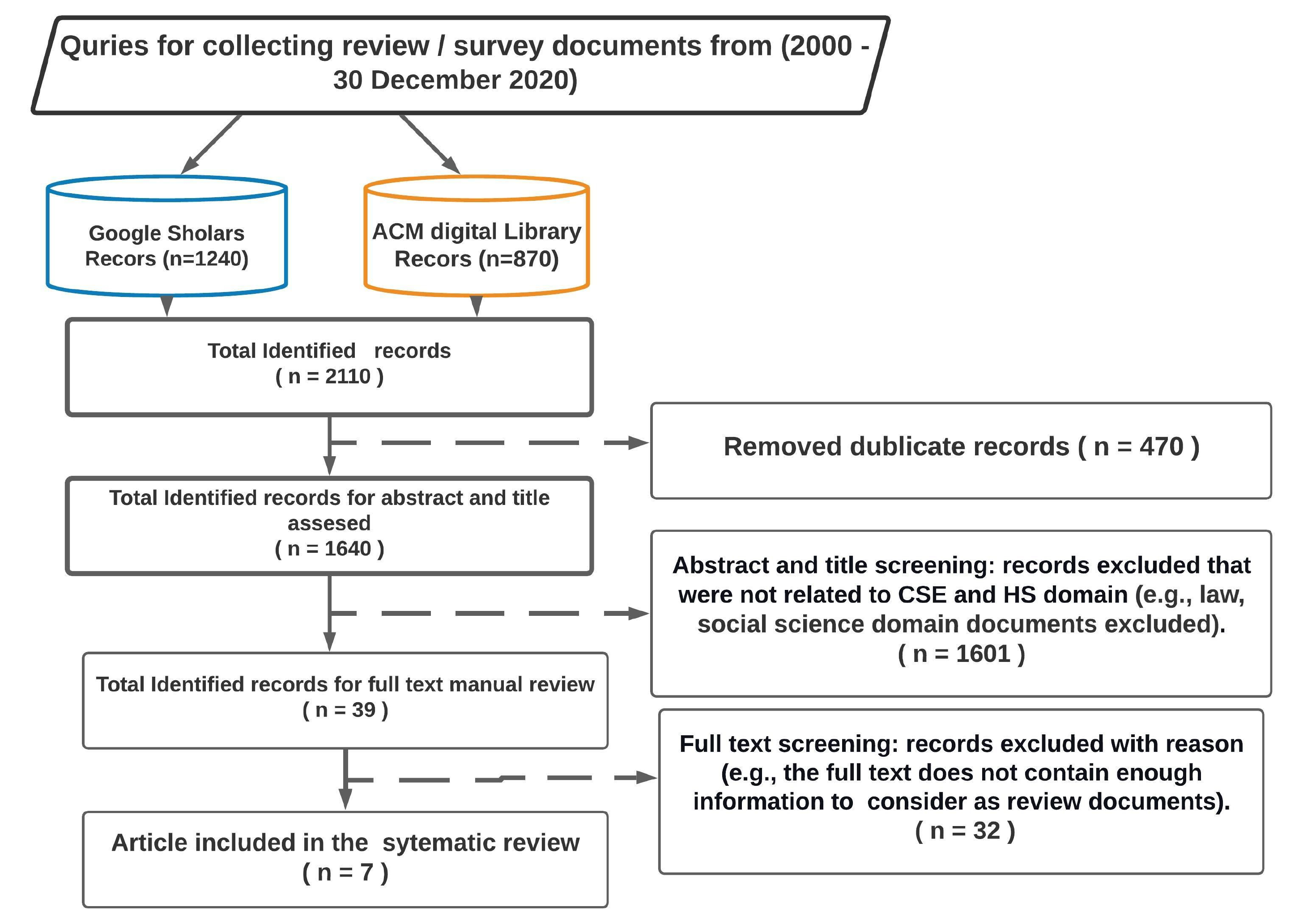}
    \caption{PRISMA flowchart for selection of previous related review/survey documents.}
    \label{fig:process_review}
\end{figure}

\subsubsection{Review of related review papers. }
The above approach identified seven review papers. The study selection process is summarized in Fig. \ref{fig:process_review}. While the initial literature search resulted 2100 records, 2061 were eliminated because either those were not review/survey documents related to HS and computer science or duplicate from both databases. The full texts of the remaining 39 reviews were carefully screened, and 32 articles were excluded because those did not have enough information to consider as review/survey documents or not related to HS/CSE domains. The remaining seven review papers, which passed the eligibility test, were divided into two main categories: narrative and systematic. Typically, narrative reviews do not reveal the methodology of data collection, in contrast to systematic reviews.

The survey of \citeauthor{Schmidt}  \cite{Schmidt} is the most cited one (more than 500 citations) \footnote{All citation mention in this papers are last searched on 03-March-20201}. The paper follows a narrative method of analysis without revealing the data collection approach. The authors provided a short, comprehensive, structured, and critical overview of the field of automatic HS detection in NLP, highlighting key terminology and focusing on feature engineering relevant to HS / bullying identification, and finally, reviewing the current dataset and societal challenges.

\begin{table}
  \caption{Review of identified survey papers on automatic hate speech detection}
  \label{Tab:Labelling example}
  \centering
  \small
\begin{tabular}{|p{4cm}|p{2cm}|p{1.3cm}|p{4cm}|l|l|}
\hline
\bf{Paper Title} &  \bf{Authors, Year} & \bf{Publisher Name} & \bf{Review Focus}& \bf{Citation} & \bf{Review Type} \\ \hline 
A Survey on Hate Speech Detection using Natural Language Processing & \citet{Schmidt} \citeyear{Schmidt}  & ACL & Features for HS Detection, anticipating alarming societal Changes, sata annotation, classification methods, etc. & 444 & Narrative\\ \hline

A Survey on Automatic Detection of Hate Speech in Text & \citet{Fortuna} \citeyear{Fortuna} & ACM & How HS work evolved from past, definition of HS, and classification method. & 211 & Systematic\\ \hline

Cyberbullies in Twitter A focused review &  \cite{Tsapatsoulis} & IEEE & Mainly for the cyberbullying on Twitter; emphasis was given to identifying Twitter abusers. & 0 & Narrative\\ \hline

DETECTION OF HATE SPEECH IN SOCIAL NETWORKS: A SURVEY ON MULTILINGUAL
CORPUS &  \citet{Al-Hassan} \citeyear{Al-Hassan}  & COSIT & Different category HS detection, multilingual HS detection, and  HS in Arabic language. & 26 & Narrative\\ \hline

Tackling Online Abuse: A Survey of Automated Abuse Detection Methods &  \citet{Mishra} \citeyear{Mishra} & ACL & Describe the existing datasets and review the computational approaches of HS detection. & 9 & Narrative\\ \hline

Resources and benchmark corpora for hate speech
detection: a systematic review &  \citet{Poletto} \citeyear{Poletto} &   Springer & Primarily focused on HS datasets. & 2 & Systematic\\ \hline

A Review on Offensive Language Detection & \citet{Pradhan} \citeyear{Pradhan}   &  Springer & Finding best strategies for HS detection. & 0 & Narrative\\ \hline

\end{tabular}
\vspace{8mm}
\end{table}

\citeauthor{Fortuna} \cite{Fortuna} is the second most cited review paper and followed a systematic review-based approach. The authors presented a critical overview of how the textual automatic detection of hate speech evolved over the past few years.  Their analysis proposed a unified and a clearer definition of the HS concept that can help build a model for the automatic detection of HS from machine learning perspective. Additionally, a comparison of performance of various HS detection algorithms is reported. Especially, they found that due to the lack of standards in the dataset, which could make the result biased. We found this review in term of its approach very relevant to our study. However, since this review was conducted at the end of 2017, an update literature is needed.  

The survey in \citeauthor{Tsapatsoulis} \cite{Tsapatsoulis} focused on cyberbullying on Twitter where an emphasis was given on identifying Twitter abusers and indicated steps required to develop practical applications for the detection of Cyberbullers.

\citeauthor{Al-Hassan}  \cite{Al-Hassan} presented a summary of several discussed papers, organized according to their publication date. Their review covered i) English Anti-social behaviors; ii) English hate speech, and finally, iii) Arabic Anti-social behaviors. The provided tables can serve as a quick reference for all the key works performed on automatic HS detection on social media. The approaches and their respective experimental results were listed concisely. They also provided a summary of multilingual contributions directly related to hate speech, with a special focus to the Arabic language in social media platforms.
\citeauthor{Mishra} \cite{Mishra} examined the existing HS datasets and reviewed the computational approaches to abuse detection, analyzing their strengths and weaknesses, discussing the emerging prominent trends, while highlighting the remaining challenges and outlining possible solutions.

\citeauthor{Poletto}  \cite{Poletto}  followed a systematic review approach to analyze existing HS resources including their development methodology, topical focus, language coverage, and other factors. The results of their analysis highlighted a heterogeneous, growing landscape marked by several issues and room for improvement.

\citeauthor{Pradhan} \cite{Pradhan} reviewed strategies for tackling the problem of identifying offense, aggression, and hate speech in user's textual posts as well as comments, micro-blogs, from Twitter and Wikipedia. Although the coverage of the dataset investigated was quite narrow. 

Table \ref{Tab:Labelling example} shows the result of previous reviews study. Only two of the identified reviews were found to follow a systematic-review based approach (\cite{Fortuna} and \cite{Poletto}) in this field. Unfortunately, one of the review \citeauthor{Fortuna} was performed in late 2017 does not cover more recent work in the field. On the other hand, the work by \citeauthor{Poletto} is mainly focused on HS datasets corpora. The rest of the identified review papers were not systematic. Some of them reviewed a small number of documents or targeted a specific part of this field (e.g., \citeauthor{Tsapatsoulis} focused on Twitter cyberbullying). The lack of coverage in the aforementioned reviews and the need for literature up-to-date in the existing systematic reviews, together with a focus on the machine and learning perspective, were the main motivation for the current review work, which aims to fill in this gap. 

Our approach is complementary to that of \cite{Fortuna} in the sense that it follows a systematic review based approach as well but it presents a more up-to-date literature and bears also some key differences. First, we adopt PRISMA protocol for systematic review-based analysis, with two search databases (Google scholar and ACM digital library). Second, a special care during the scrutinizing phase is devoted to track machine learning and deep learning methods with their associated evaluation performance and dataset employed. Third, we track the timely evolution of records and methods. We also enumerate existing data collections, including multilingual more comprehensively than in previous studies. Fourth, we summarize existing open-source projects and valuable resources in the field, and finally, we highlight the key challenges and open research agenda. 

\section{Systematic literature review methodology for collecting  hate speech documents}

\label{METHODOLOGY}
The previous section~\ref{Related_work} presented a brief systematic review of past review papers related to HS automatic detection. This section details our systematic literature review regarding HS automatic detection. For this purpose, we adopted PRISMA framework as in \citeauthor{moher2009preferred}, highlighting the keyword selection, Search sources, and filtering process. 

\begin{figure}
    \centering
    \includegraphics[width=.8\linewidth,]{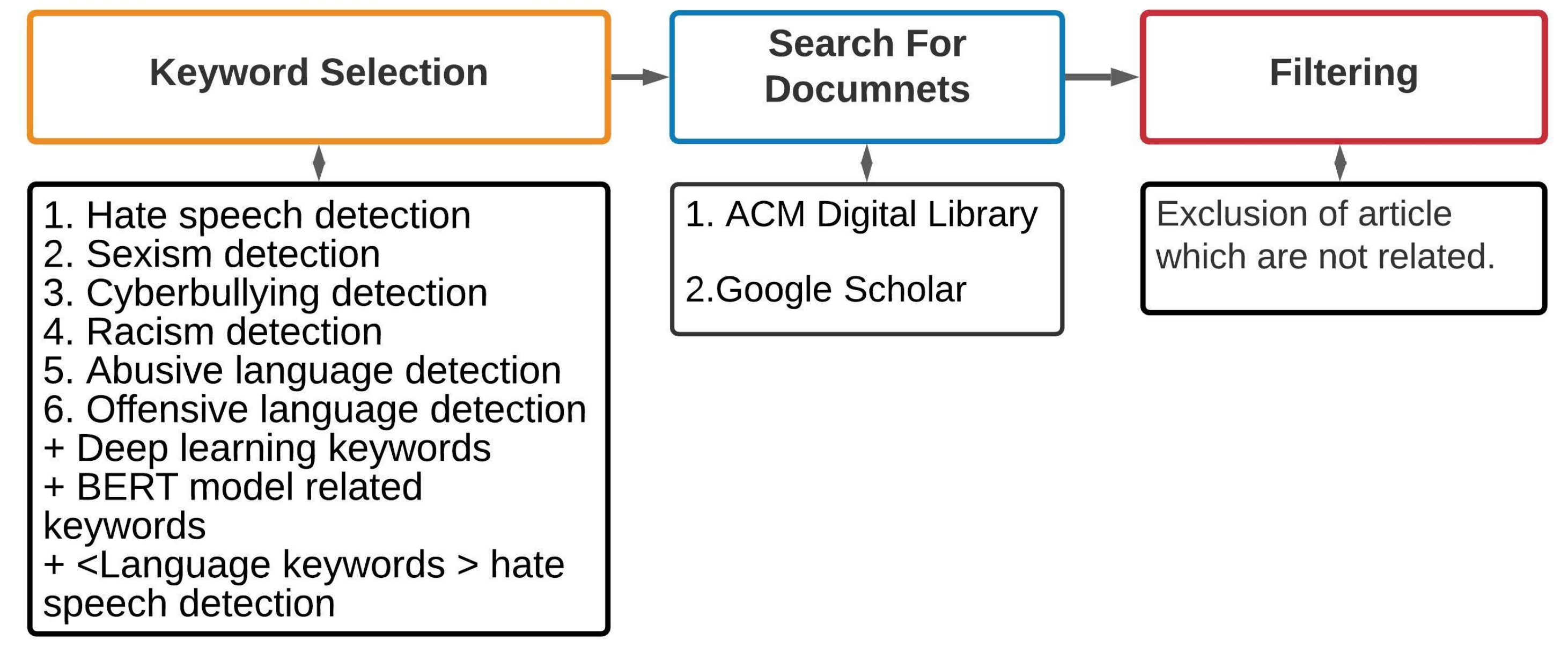}
    \caption{Methodology for document collection.}
    \label{fig:process}
\end{figure}

\subsection{Keyword selection}
The first phase conducted was the keywords selection. Since hate speech is a concept that comprises broad hate categories, our search criteria were partitioned into six categories: hate speech, sexism, racism, cyberbullying, abusive, and offensive. This provides us the best chance of retrieving a significant number of relevant work. Besides, as we wanted to pay special attention to machine learning and deep learning-based methods, several related abbreviations and keywords have been accommodated and added to the keyword search (i.e., CNN, LSTM, RNN, BERT, etc.).  Furthermore, 20 top-speaking languages \footnote{The Ethnologue \url{https://www.ethnologue.com/}, one of the trusted platforms regarding language information based on more than a thousand bibliography references.}  added in search keywords to retrieve multilingual works. A selected list of keywords is shown in Table \ref{tab:KeyWord}:  

\begin{table}
    \centering
    \caption{Keywords list for the query search.} 
    \label{tab:KeyWord} 
    \scalebox{.85}{\begin{tabular}{|p{4cm}|p{5cm}|p{4cm}|p{4cm}|}
    \hline
    \bf{Different type HS   keywords} &  \bf{Deep learning Keywords} &  \bf{Deep learning BERT model related  keywords}  &  \bf{Language  keywords}  \\ \hline
    
\begin{enumerate}
    \item Hate speech detection
    \item Sexism detection
    \item Cyberbullying detection
    \item Racism detection
    \item Abusive language detection
    \item Offensive language detection
\end{enumerate} &

\begin{enumerate}
    \item Deep learning hate speech
    \item CNN hate speech  
     \item LSTM hate speech  
     \item RNN hate speech  
     \item Deep learning Cyberbullying  
     \item CNN Cyberbullying  
     \item LSTM Cyberbullying  
     \item Deep Learning Offensive  
     \item CNN Offensive  
     \item LSTM Offensive \end{enumerate} & 
     
      \begin{enumerate}
    \item BERT hate speech 
     \item BERT Cyberbullying  
     \item BERT Abusive  
     \item BERT Rasism
     \item BERT Sexism 
     \item BERT Offensive   
 \end{enumerate} & Chinese (Mandaren, Yue, Wu), Hindi, Spanish, Arabic, Bengali, French, Russian, Portuguese, Urdu, Indonesian, German, Japanese, Marathi, Telugu, Turkish, Tamil, Korean.
 
 \\\hline
   \end{tabular}}
\end{table}

\clearpage

\subsection{Search sources} We used two different databases  (ACM Digital Library and  Google Scholar), aiming to gather the most significant number of records in the areas of computer science and engineering (CSE). This is motivated by the availability of search through an API, allowing the application of simple NLP modules to identify duplication and check string matching as well as record statistical trends. On the other hand, our desire to focus on computer science aspect of HS detection makes ACM library  as an ideal candidate for search database, while Google scholar expects to identify all other relevant and high impact results outside ACM community.  

The input database search consists of an OR-logical combination of keywords of individual categories (hate speech OR sexism OR racism... etc.), see Table \ref{tab:KeyWord} for detailed listing of such keywords. Besides, this process has been automated by utilizing the beautiful-soup python web crawler for both databases (ACM and Google Scholar) by monitoring the API output, which consists of the paper's title, publication year,  abstract, author's name, publisher information, citation, and link to the full article. We have made our scraping code publicly available for the community \footnote{\url{https://github.com/saroarjahan/Google_sholars_ACM_digital_library_crawler}}. Initially, we collected papers from 2000 to 2021, and no language restrictions were imposed since we wanted to gather multilingual work associated with hate speech detection. The last search for article collection was run on 18 March 2021.

The title, publication year, abstract, author's names, publication venue, and link of full papers were exported to an MS Excel spreadsheet for further analysis.

\begin{figure}
    \centering
    \includegraphics[width=.75\linewidth,]{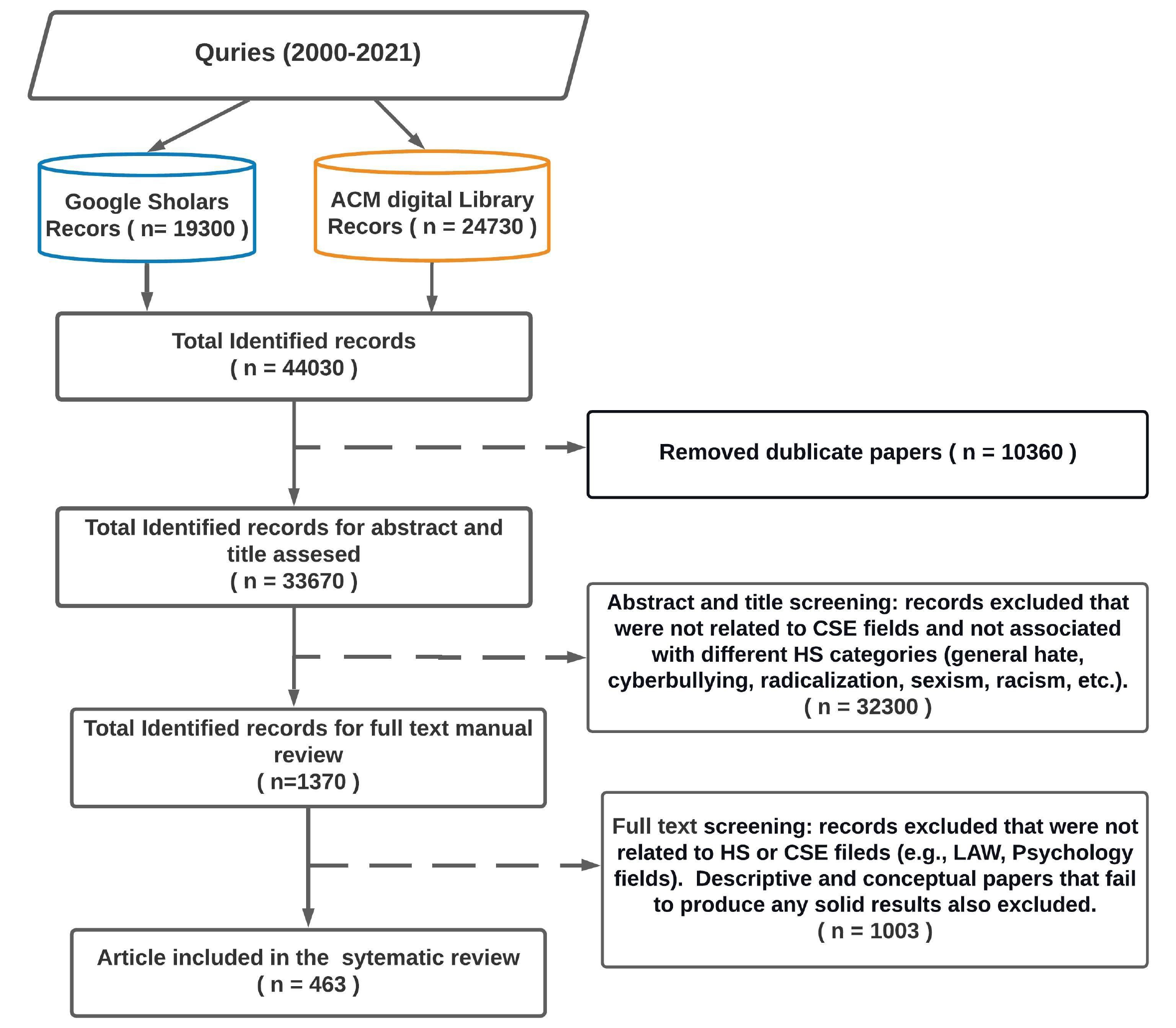}
    \caption{PRISMA flowchart for the selected studies for systematic review.}
    \label{fig:prisma}
\end{figure}
\clearpage

\subsection{Filtering Documents}
The PRISMA flow chart highlighting the inclusion and exclusion criteria for the document search and inclusion in the database is summarized in Figure \ref{fig:prisma}. Initially, 44,030 documents were collected from  2000 to 2021. Since we have collected data from two different databases, duplicate papers have been removed automatically from the system, leaving 33670 records for further title and abstract scrutinizing. In this respect, most documents that were not related to CSE fields and not associated with different hate speech categories (general hate, cyberbullying, abusive, offensive, sexism, racism, etc.) were excluded after the title and abstract screening.

The remaining  1329 papers were considered for full-text review. Two independent reviewers with knowledge in this field have carefully performed this manual scrutinizing task. Those articles that were not related to hate speech or CSE fields (e.g., LAW, Psychology  fields)  were discarded. Similarly, descriptive and conceptual papers that fail to produce any solid results were also ignored. During this phase, disagreements between the two reviewers were discussed and resolved by consensus. If no agreement could be reached, the views of a third reviewer would have been taken into consideration. 

Finally, 463 articles were considered for the final systematic review and analysis. Among those 463 papers, 96 papers have been found to follow deep learning methods.

\section{Systematic review results}
\label{results}
\subsection{Number of publications per year}
\begin{figure}
    \centering
    \includegraphics[width=.8\linewidth,]{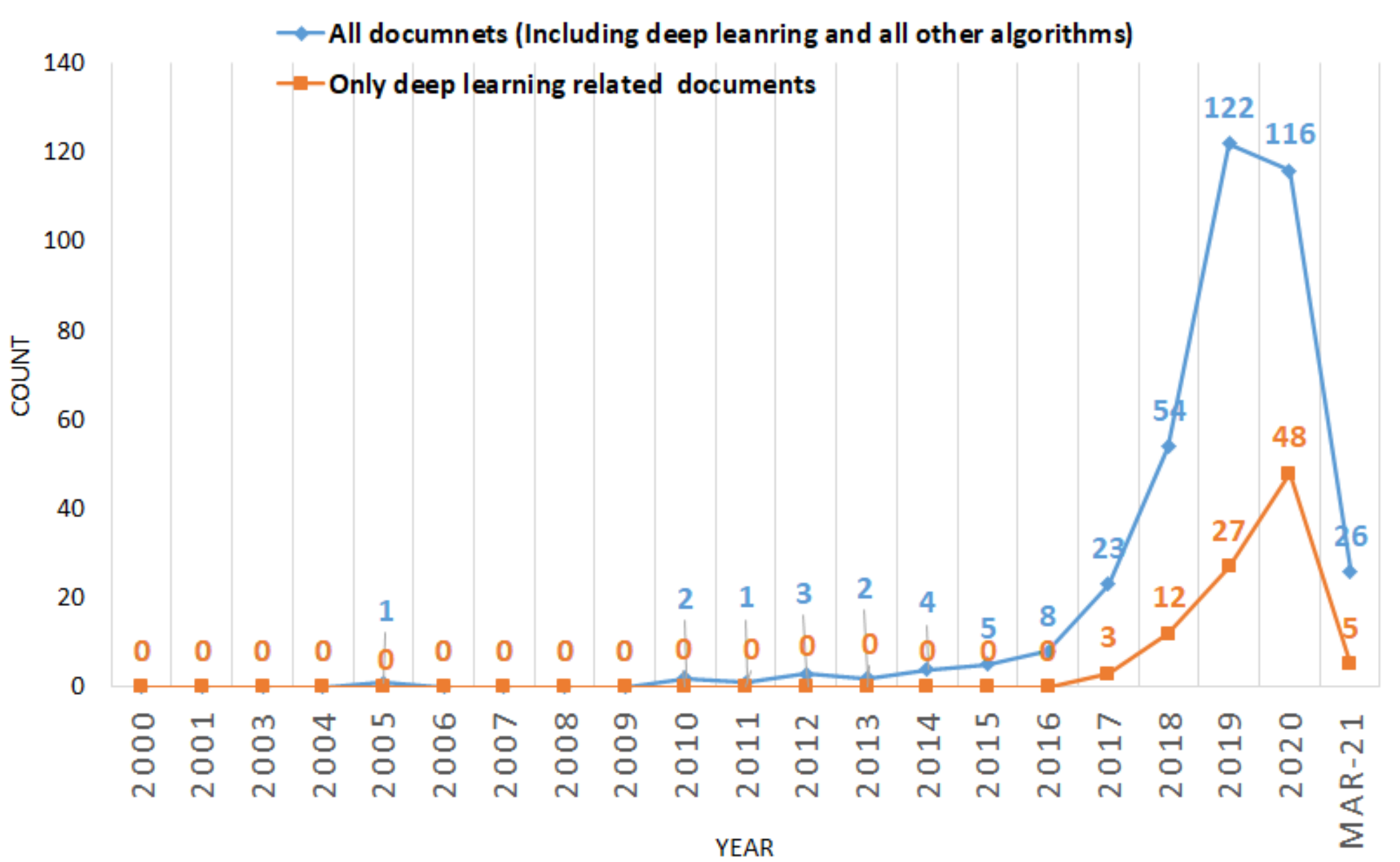}
    \caption{Number of publications per year from 2000-2021 related automatic hate speech detection in NLP (blue line represent all 463 documents  including deep learning and other ML approach, and yellow  line represent 96 documents related to Deep-learning method).}
    \label{fig:numberpaper}
\end{figure}

As we can see in Figure \ref{fig:numberpaper}, a total of 463 papers were identified from 2000 to 2021 (including deep learning and all other methods). Before 2010, we have found only 1 document related to hate speech. From 2010 to 2016, only 25 papers associated with HS detection were found, yet there was no work related to deep learning. However, since 2017 the number of published documents raised rapidly with a steady increase of deep learning based HS detection approach. A total of 96 documents were found from 2017 to 2021 using deep-learning HS detection, indicating a trend of almost doubling the number of deep learning approach each year. The relatively small value in 2021 is due to the fact that the collection of new documents stopped in March 18, 2021.

\clearpage

\subsection{Publication Venue}
\begin{figure}
    \centering
    \includegraphics[width=.8\linewidth,]{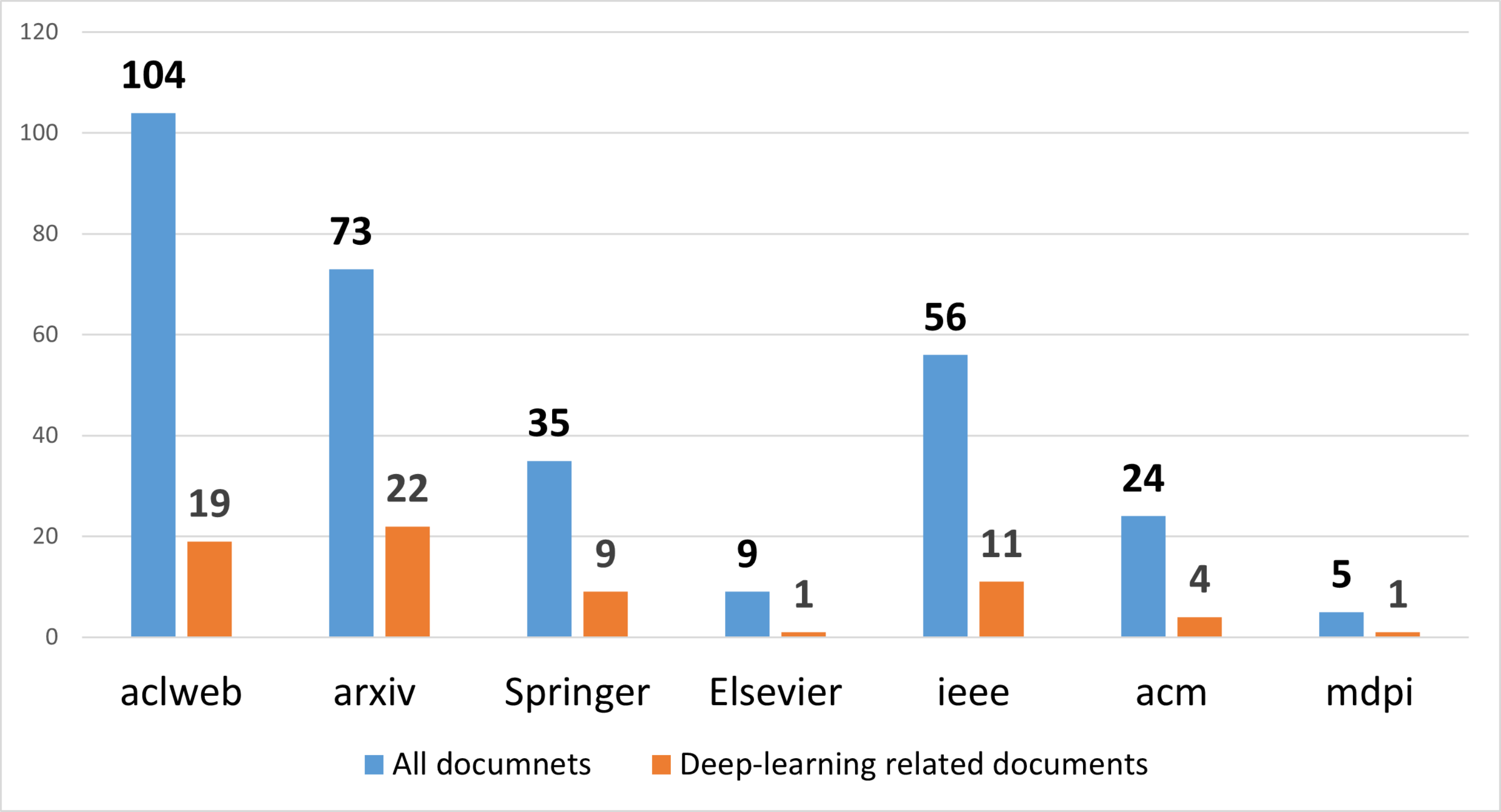}
    \caption{Most Used Data Platforms (blue histogram represent all 463 documents  including deep learning and other ML approach, and yellow histogram represent 96 documents related to Deep-learning method).}
    \label{fig:venue}
\end{figure}

We have scrutinized the obtained records with respect to publication venues in an attempt to identify any dominating trend. From the total of 463 identified documents in textual HS automatic detection, we have found 72 different venues. The publication venues with more than 4 occurrences in our collection are presented in Figure~\ref{fig:venue}. The most common platforms for publication of hate speech documents were ACLWEB\footnote{\url{https://www.aclweb.org/portal/}}, ArXiv\footnote{\url{https://arxiv.org/}}, IEEE\footnote{\url{https://www.ieee.org/}}, Springer\footnote{\url{https://www.springer.com/gp}}, and ACM\footnote{\url{https://www.acm.org/}}. The Association for Computational Linguistics (ACL) is the premier international scientific and professional society working on NLP's computational problems. Therefore, a vast number of papers were published in ACL-WEB forums. The second most popular source was Arvix, an open-access repository of electronic preprints. This can partially be explained by the fact that the hate speech detection area has become popular, with many autonomous and exploratory work being conducted. Additionally, this high number of publication venues reveals that HS automatic detection is not limited to a few sources of publication venue and testifies the field's multidisciplinary nature.

\clearpage

\subsection{Categorization according to methods employed and detection performance}
This section first reviews the results (identified documents) in terms of machine learning and feature employed, the platform used for dataset collection, category of hate speech investigated, and bibliometric data of the publication\footnote{All citations last updated on 03-Mar-2021} as well as performance metrics (Accuracy (Acc), Precision (P), Recall (R)) claimed by the authors. After that, we evaluated the state-of-the-art of deep-learning methods related to HS automatic detection.

\subsubsection{Statistical trends of results} 
Tables \ref{tab:Englishdata} and \ref{tab:multilanguagedata} present some characteristics of selected highly cited papers, organized according to their publication date. The tables can serve as a quick reference for all the key works performed in textual automatic HS detection. The employed methods and related experimental results in terms of Precision, Recall, or F-measure are listed concisely. Table \ref{tab:Englishdata}'s attributes include data platform, type of hate speech, type of machine learning, features, and performance results. One notices, for instance, the dominance of Twitter and YouTube as the main source of data and supervised machine learning algorithms as the dominant machine learning type. Whereas Table \ref{tab:multilanguagedata} repeats the previous process for non-English and multilingual textual data. Furthermore, we have plotted statistical figures regarding all the 463 documents to understand the overall trend according to language, types of HS, data collection platform, ML approach, features, and algorithm used for HS detection.

\begin{figure}
    \centering
    \includegraphics[width=.6\linewidth,]{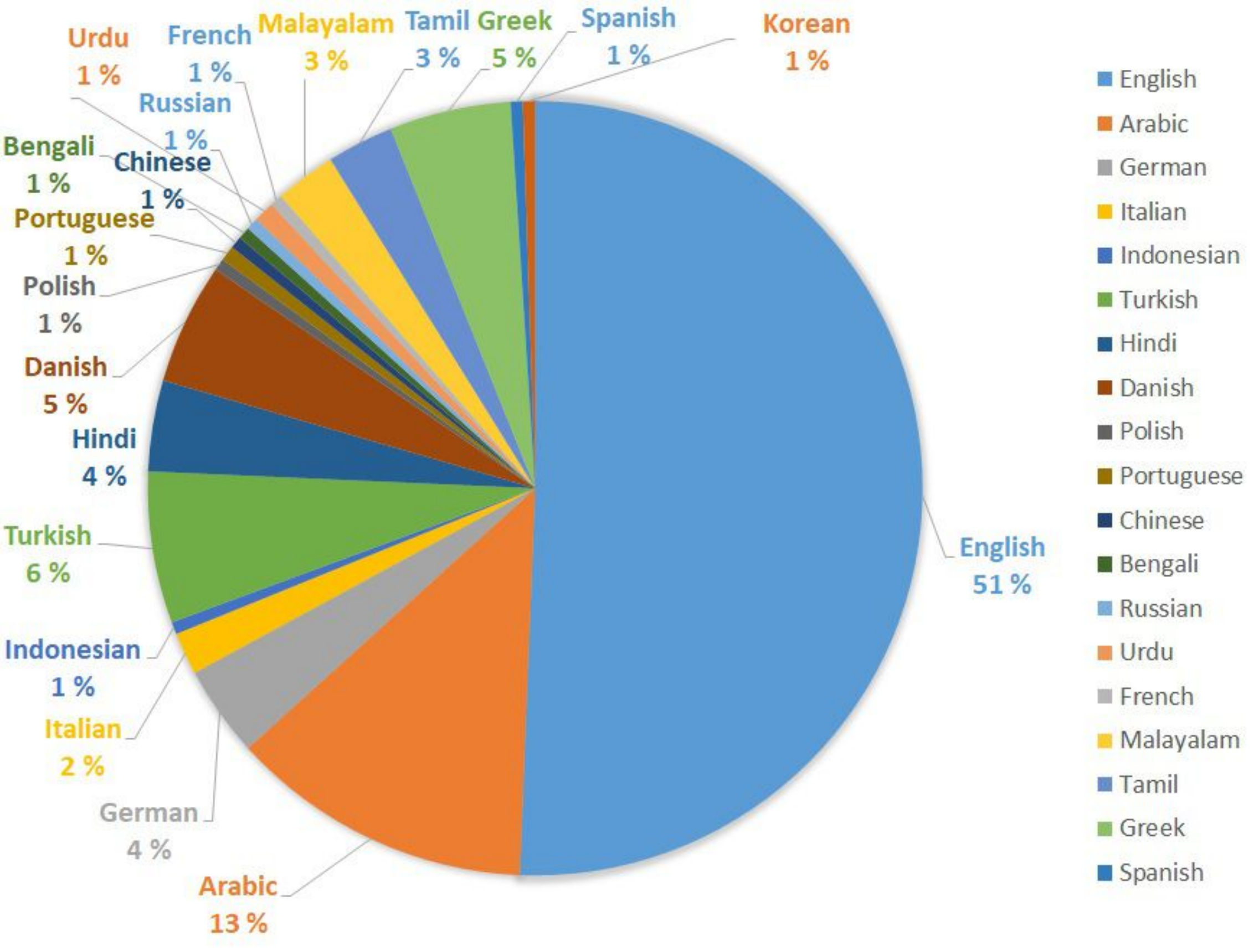}
    \caption{Statistics of the percentage of previous HS wrok in different language.}
    \label{fig:statistic_language}
\end{figure}

Figure \ref{fig:statistic_language} illustrates the proportion of the various languages of the textual input analyzed by HS automatic detection algorithms in the identified records. As expected, English textual source is by far the most investigated. This is rationally justified by the fact that initial work in the field has started with English textual dataset as in  \citeauthor{dinakar2012common} \cite{dinakar2012common}, but also due to the maturity of the natural language Parsers developed for  English language as well as the abundance of benchmarking dataset that enable researchers to carry out useful comparative results. Nevertheless, with the advances in multilingual parsers and deep learning technology, together with increasing pressures from policy-makers to handle hate speech issues at local resources, non-English HS detection toolkits have seen a steady increase. The figure indicates that about 51\% of all works in this field are performed on English dataset, with an increase of proportion of other languages as well where Arabic (13\% ) \cite{mubarak2017abusive,hassan2020alt, alami2020lisac, wang2020galileo},  Turkish (6\%) \cite{wang2020galileo, ozdemir2020nlp}, Greek (4\%) \cite{wang2020galileo, ahn2020nlpdove, socha2020ks}, Danish (5\%) \cite{pamies2020lt,wang2020galileo}, Hindi (4\%) \cite{raja2021nsit,bashar2020qutnocturnal, mishra20193idiots},  German (4\% ) \cite{kumar2020comma, quea2020simon},  Malayalam (3\%) \cite{m1,m2}, Tamil (3\%) \cite{m1,t}, Chinese (1\%) \cite{su2017rephrasing, tang2020categorizing, yang2020tocp}, Italian (2\%) \cite{polignano2019alberto}, Urdu (1\%) \cite{u1, u2, u3}, Russian(1\%) \cite{andrusyak2018detection}, Bengali (1\% ) \cite{b1, b2, karim2020classification}, Korean (1\%) \cite{k}, French (1\%) \cite{aluru2020deep, ousidhoum2019multilingual, ghanghor2021iiitk}, Indonesian (1\%) \cite{alfina2017hate}, Portuguese (1\%) \cite{alfina2017hate}, Spanish (1\%) \cite{gonzalez2021twilbert} and  Polish (1\%) \cite{ptaszynski2019results} seem to dominate the rest of the languages in this field.

It is worth noticing that despite Chinese being the 2nd largest speaking language globally, it has been much less investigated in HS detection community. One reason could be the lack of Chinese language in HS competition workshop such as SemEval2020 \cite{zampieri2020semeval} and Hasoc2020 \cite{mandl2020overview} where multilingual tasks included English, Danish, Greek, Arabic, Tamil, Malayalam, Hindi, German and Turkish, which encouraged many participants all over the world to work in these languages.

\begin{figure}
    \centering
    \includegraphics[width=.7\linewidth,]{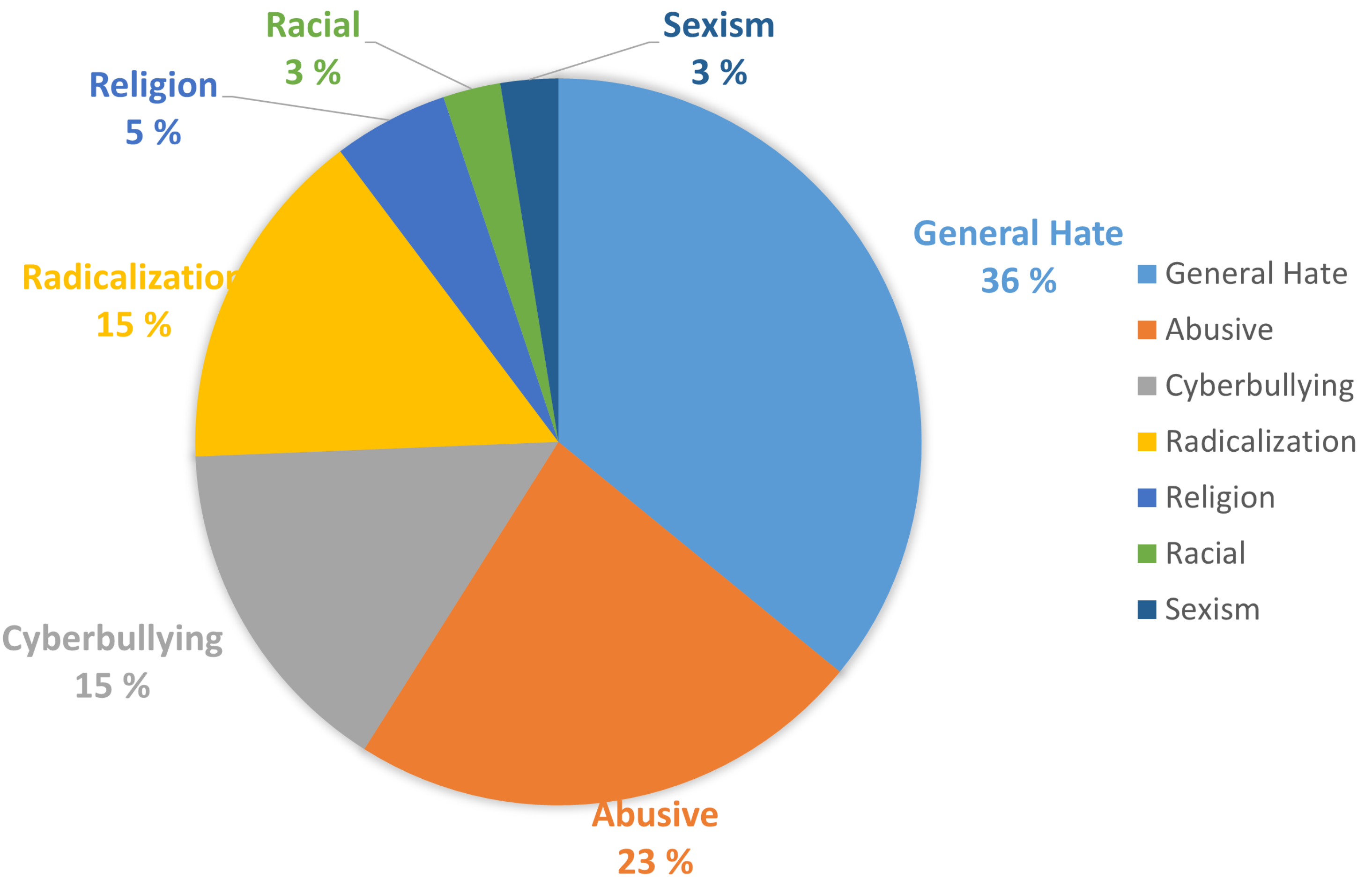}
    \caption{Statistics of the percentage of HS work in different categories.}
    \label{fig:statistic_type}
\end{figure}

Figure \ref{fig:statistic_type} depicts the percentage of different HS categories in the identified records. We can see that publications related to 'general hate' (36\%) are a dominant trend followed by 'abusive language' 23\% of total records. Cyberbullying and Radicalization categories share the same percentage of (15\%) each. While relatively a small percentage is assigned to religion (5\%), racial (3\%), and sexism (3\%) associated hate speech categories.

\begin{figure}
    \centering
    \includegraphics[width=.7\linewidth,]{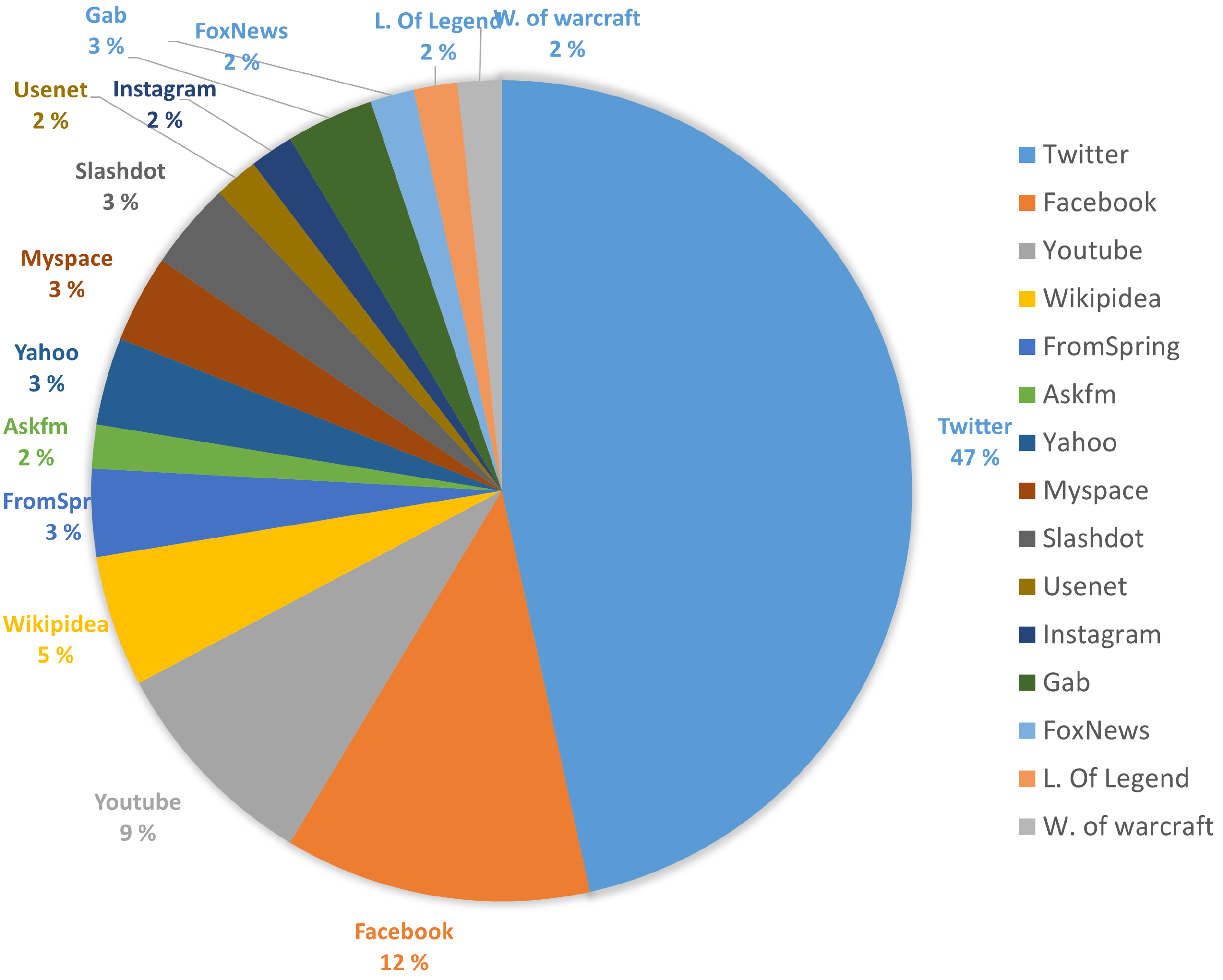}
    \caption{Statistics of platforms used for data collection.}
    \label{fig:statistic_platform}
\end{figure}

From Figure~\ref{fig:statistic_platform}, one can notice that 47\% of dataset employed in HS detection was collected from Twitter social network, followed by Facebook (12\%), Youtube (9\%) and Wikipedia (5\%). This indicates a growing trend in research community to tackle the occurrence of hate speech in social media forums as social media platforms constitute by far the dominant agora of hate speech because of easy access, fast spread and societal impact. The rest of the dataset sources, with low occurrence rate, were mainly investigated for comparative  purpose and benchmarking. For instance, \citeauthor{pawar2018cyberbullying} used Formspring dataset for cyberbullying detection and developed methods that enhance question-answer systems. In Section~\ref{resources}, we discuss in detail the current state of the art of HS dataset.

\begin{figure}
    \centering
    \includegraphics[width=.7\linewidth,]{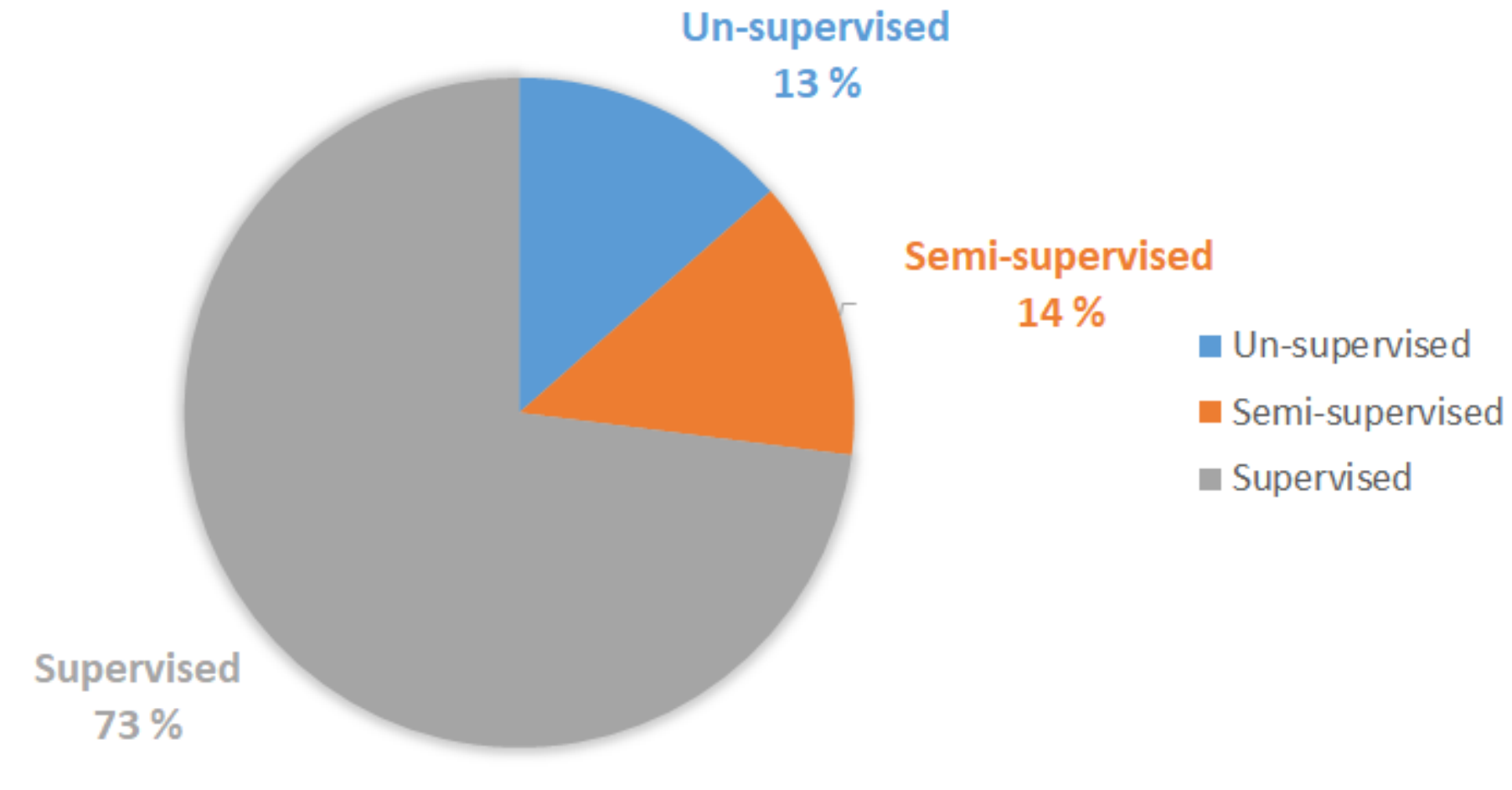}
    \caption{Statistics on types of ML approach used for HS detection ( e.g., supervised, semi-supervised or unsupervised).}
    \label{fig:statistic_ML}
\end{figure}

Figure~\ref{fig:statistic_ML} provides a global trend in terms of types of machine learning approaches employed in our identified records. Among the ML approaches, we distinguish supervised, semi-supervised, and unsupervised like approaches. The analysis revealed that most of the works adopted supervised methods (73\%). From Tables \ref{tab:Englishdata} and \ref{tab:multilanguagedata}, we can observe that any of these three methods can achieve high-performance accuracy, and there is no substantial evidence to favor one over another one, whereas only the context of data (e.g., availability and quality of training samples) can play a role in deciding about the suitability of one category over another one. For example, \citeauthor{chen2012detecting} used an unsupervised method and lexical \& syntactic features to achieve 98\% accuracy. Similarly, several works were based on supervised and semi-supervised methods that have shown close or better performance \cite{badjatiya2017deep} \cite{abozinadah2016improved}. Nevertheless, it is worth mentioning the popularity of the supervised like approach over other ML approaches, possibility due to the multiplication of benchmarking dataset and machine learning / deep learning platforms that promote supervised approach.

\begin{figure}
    \centering
    \includegraphics[width=.7\linewidth,]{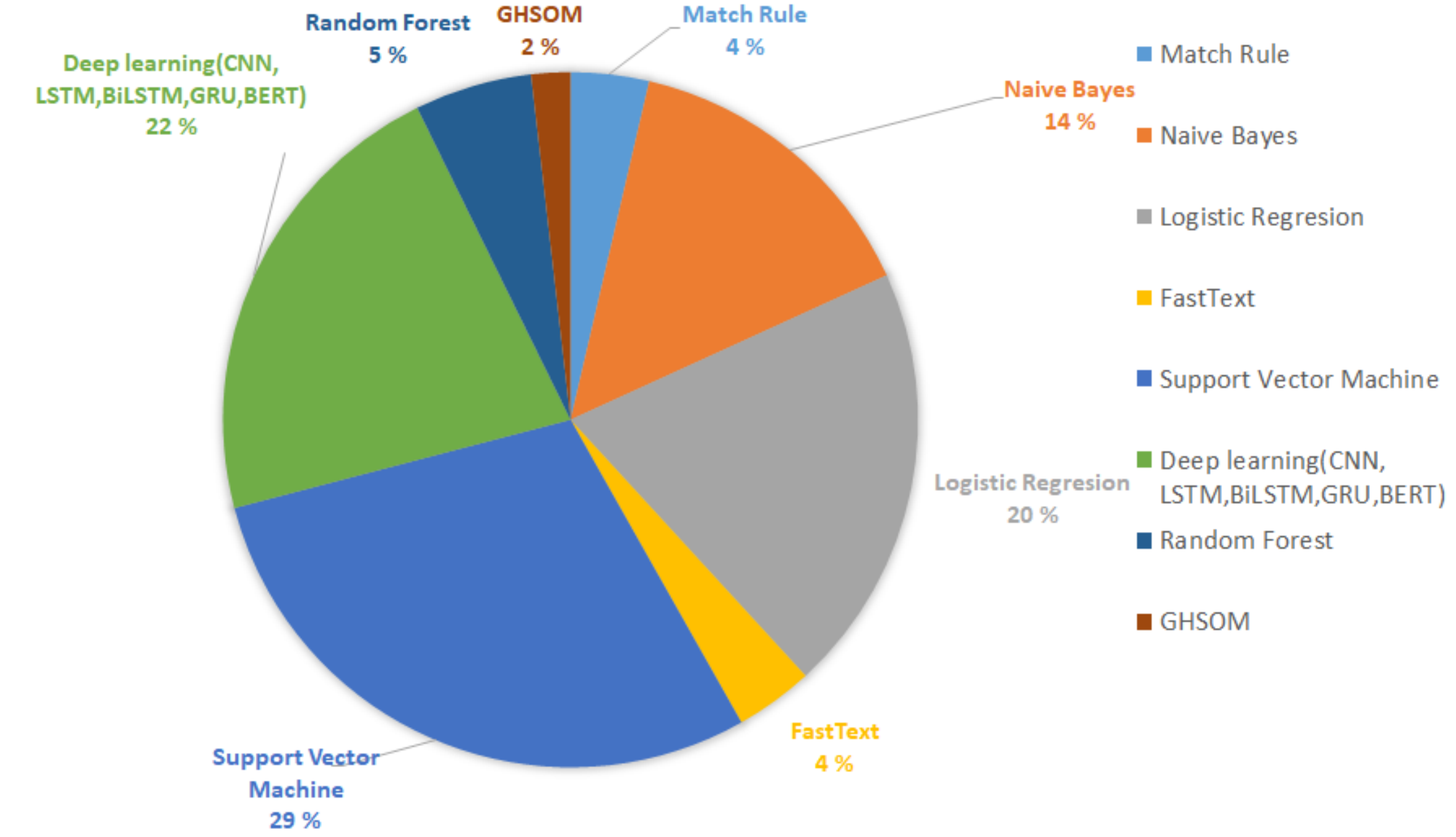}
    \caption{Statistics of algorithm types used for HS detection.}
    \label{fig:statistic_all_algorithm}
\end{figure}

\begin{figure}
    \centering
    
    \includegraphics[width=.8\linewidth,]{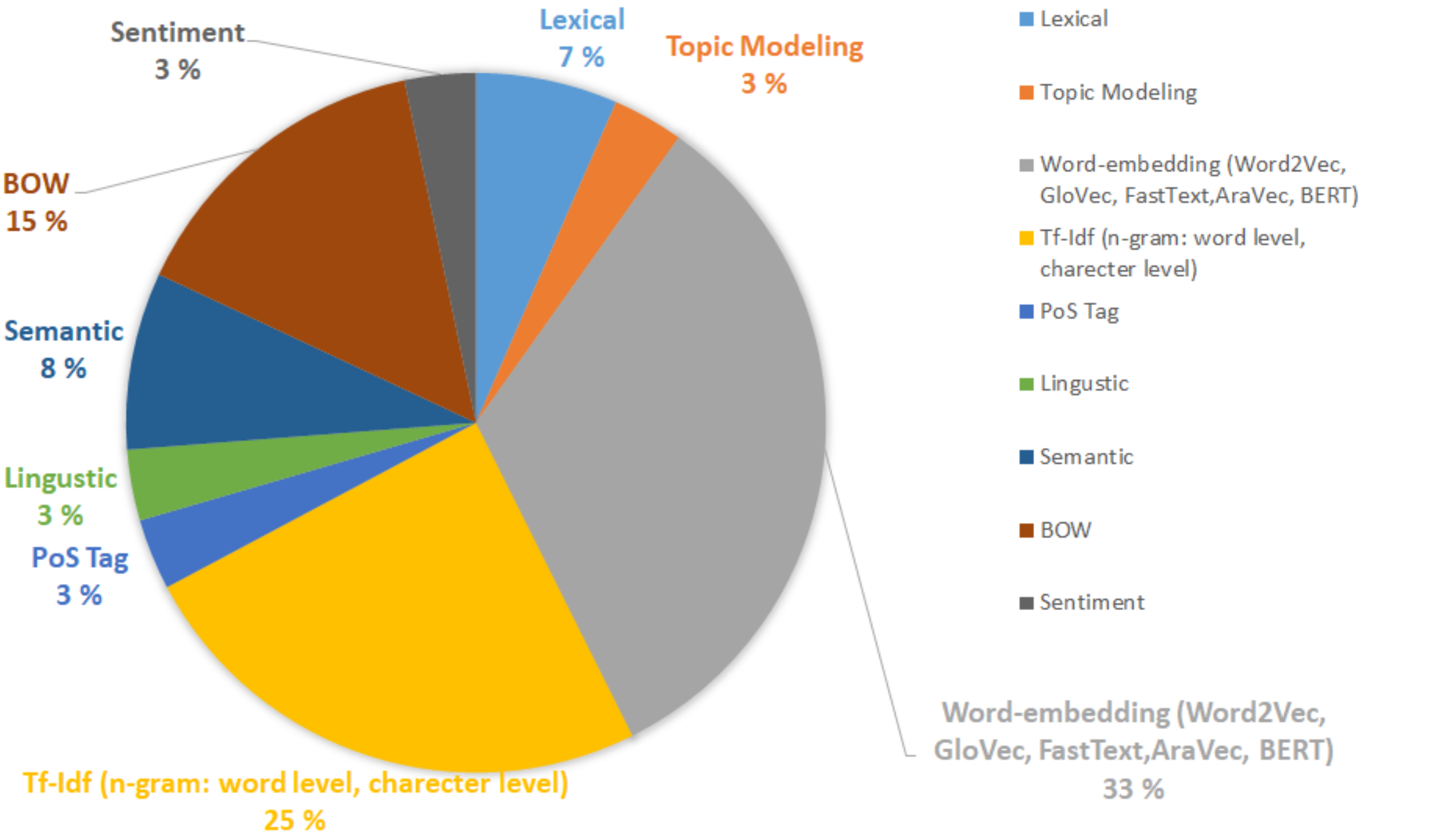}
    \caption{Statistics of features employed by the ML/deep learning algorithms.}
    
    \label{fig:statistic_all_feature}
\end{figure}

Fig.~\ref{fig:statistic_all_algorithm} and \ref{fig:statistic_all_feature} depict the percentage of ML algorithms and features employed in the identified records, respectively. The SVM method emerges as the most popular HS detection model covering 29\% of total records. The use of deep learning models to HS started to rise from 2017 \cite{badjatiya2017deep}\cite{park2017one} to quickly cover about 22\% of total identified records. On the other hand, LR (20\%) and  NB (14\%) were also among popular ML methods investigated by the researchers. We also noticed that in many deep-learning related methods, non-deep learning models were often employed as baseline to compare the performance of the investigated deep-learning model \cite{dowlagar2021hasocone} \cite{al2021detection} \cite{jahan2020team} \cite{badjatiya2017deep}.

Fig.~\ref{fig:statistic_all_feature} shows that TF-IDF based features cover 29\% of the total records. However, word embedding models, which have widely being used in deep learning embedding layers, cover 33\% of the entire records. The PoStag (3\%), topic modeling (3\%), and sentiment (3\%) features were the least used features. This suggests that the deep-learning models  and embedding features seem comparatively popular and widely used by the community. Next, we explored the popularity of the various deep-learning architectures (e.g., CNN, LSTM, BiLSTM, etc.) for HS automatic detection. The results are outlined in the next section \ref{Deep-learning records}.

\clearpage

\begin{table}
    \centering
    \caption{Summary of key contributions for English HS detection and their performance in terms of Precision (P), Recall (R), F1-Score (F), Citation (C)} 
    \label{tab:Englishdata} 
    \scalebox{.75}{\begin{tabular}{|p{2.5cm}|p{2.3cm}|p{2.2cm}|p{2cm} |p{4cm}|p{2.3cm}|l|l|l|l|}
    \hline
    \bf{Author, Year} &  \bf{Platform} & \bf{Type} & \bf{ML Approach}& \bf{Features Representation} & \bf{Algorithm}  & \bf{P}& \bf{R} & \bf{F}& \bf{C} \\ \hline

\citet{chen2012detecting} \citeyear{chen2012detecting} & YouTube &   Abusive & Un-Supervised & Lexical and syntactic  & Match Rules & 0.98 & 0.94 &- &472 \\\hline

\citet{xiang2012detecting} \citeyear{xiang2012detecting} & Twitter &   Abusive & Semi-Supervised & Topic modelling & LR & - & - & 0.84 & 213 \\\hline

\citet{dinakar2012common} \citeyear{dinakar2012common}& YouTube  & Cyberbullying & Supervised  & Tf-idf, lexicon, PoS tag, bigram & SVM& .66 & - &- & 446 \\\hline

\citet{warner2012detecting} \citeyear{warner2012detecting} & Yahoo, news-group & Radicalization  & Supervised & Template-based, PoS tagging & SVM & .59 & .68 & .63 & 432 \\\hline

\citet{wadhwa2013tracking} \citeyear{wadhwa2013tracking} & Twitter & Radicalization & Un-Supervised & Topic identification, N-grams &  Topic-entity mapping & - & -& -& 29\\\hline

\citet{kwok2013locate} \citeyear{kwok2013locate} &  Twitter & Racism  & Supervised & Unigram & Naïve Bayes & - & - &- & 273\\\hline

\citet{nahar2014semi} \citeyear{nahar2014semi} & Myspace,
Slashdot & Cyberbullying  & Semi-Supervised & Linguistic features & Fuzzy SVM & .69 & .82 &.44 & 60\\\hline

\citet{burnap2014hate} \citeyear{burnap2014hate}& Twitter & Hate & Supervised & BOW, Dependencies, Hateful Terms & Bayesian LR & .89 &  .69 &  .77  &80\\\hline

\citet{agarwal2015using} \citeyear{agarwal2015using} & Twitter & Radicalization  & Semi-Supervised  & Linguistic,Term Frequency  & KNN, SVM  & -  & -  & .83 & 97\\\hline

\citet{gitari2015lexicon} \citeyear{gitari2015lexicon}& Blog & Hate, Weakly hate, Strongly hate & Semi-Supervised & Lexicon, Semantic, theme-based features & Rule based & 0.73 & 0.68 & 0.70 & 257 \\\hline

\citet{djuric2015hate} & Yahoo & Finance Hate, & Supervised & Paragraph2vec, & CBOW, LR & - & - & -& 409  \\\hline

\citet{waseem2016hateful} \citeyear{waseem2016hateful}& Twitter & Hate& Supervised & Character n-grams & LR & 0.72 & 0.77 & 0.73 & 665 \\\hline

\citet{di2016unsupervised} \citeyear{di2016unsupervised} & YouTube,Form-Spring, Twitter & Cyberbullying  & Un-Supervised & Semantic and syntactic features & GHSOM network and K-mean &  .60 & .94 & .74 & 26 \\\hline

\citet{park2017one} \citeyear{park2017one} & Twitter&   Abusive &  Supervised&  Character and Word2vec &  Hybrid CNN&  0.71&  0.75&  0.73 & 202 \\\hline

\citet{chen2017abusive} \citeyear{chen2017abusive} & Youtube, Myspace, SlashDot & Abusive &  Supervised & Word embeddings & FastText & - &0.76 & - & 9 \\\hline

\citet{badjatiya2017deep} \citeyear{badjatiya2017deep} & Twitter & Sexist, Racist & Supervised  & FastText, GloVe Random Embedding,Tf-IDF, BOW & LR, SVM CNN, LSTM and GBDT & 0.93 & 0.93 & 0.93 & 503  \\\hline

\citet{wiegand2018inducing} \citeyear{wiegand2018inducing} & Twitter, Wikipedia, UseNet & Abusive & Supervised & Lexical,linguistics and word embedding & SVM & .82 & .80 &0.81 &55 \\\hline

\citet{pawar2018cyberbullying} \citeyear{pawar2018cyberbullying}& Form-spring & Cyberbullying  & Supervised & BOW & M-NB and Stochastic Gradient Descent & - & - & .90 & 7 \\\hline

\citet{watanabe2018hate} \citeyear{watanabe2018hate}& Twitter & Hate, Offensive  & Supervised & Sentiment-Based, Semantic, Unigram & J48graft & 0.79 & 0.78 & 0.78 & 95 \\\hline

\citet{malmasi2018challenges} \citeyear{malmasi2018challenges} & Twitter& Hate, offensive& Supervised& N-grams, Skip-grams, hierarchical, word clusters& RBF kernel, SVM& 0.78& 0.80& 0.79 & 133 \\\hline

\citet{pitsilis2018effective} \citeyear{pitsilis2018effective} & Twitter & Racism or Sexism & Supervised  & Word-based frequency, vectorization & RNN and LSTM & 0.90 & 0.87 & 0.88  & 62\\\hline

\citet{fernandez2018contextual} \citeyear{fernandez2018contextual}&  Twitter & Radicalization & Supervised&  Semantic Context & SVM & .85 & .84 & .85 & 12 \\\hline

\citet{ousidhoum2019multilingual} \citeyear{ousidhoum2019multilingual} &  Twitter & Sexual orientation, Religion, Disability & Supervised & BOW & LR, biLSTM  & - & - & 94 & 35 \\\hline

\citet{zhang2019hate} \citeyear{zhang2019hate}& Twitter & Racism, Sexism & Supervised  & Word embeddings & CNN+GRU & - & - & 0.94 & 98   \\\hline

   \end{tabular}}
\end{table}

\begin{table}
    \centering
    \caption{Summary of key contributions for non-English language in HS detection.} 
    \label{tab:multilanguagedata} 
    \scalebox{.75}{\begin{tabular}{|p{2.5cm}|p{1.5cm}|p{1.3cm}|p{1.7cm}|p{2cm} |p{4cm}|p{2.3cm}|p{.8cm}|p{.6cm}|l|l|}
    \hline
    \bf{Author, Year} &  \bf{Platform} & \bf{Language} & \bf{Class} & \bf{ML Approach}& \bf{Features Representation} & \bf{Algorithm}  & \bf{P}& \bf{R} & \bf{F}& \bf{C} \\ \hline
    
\citet{abozinadah2015detection} \citeyear{abozinadah2015detection}& Twitter & Arabic & Abusive  & Supervised & Profile and tweet-based features, bag of words, N-gram, TF-IDF & Naïve Bayes & 0.85 & 0.85 & 0.85 & 40 \\ \hline

\citet{magdy2015failedrevolutions} \citeyear{magdy2015failedrevolutions}& Twitter  & Arabic  & Terrorism (Pro-ISIS and Anti-ISIS) & Supervised & Temporal patterns, Hashtags & SVM & 0.87 & 0.87 & 0.87 & 101\\ \hline

\citet{kaati2015detecting} \citeyear{kaati2015detecting} & Twitter  & Arabic & Terrorism (Support or Oppose Jihadism) & Semi-Supervised & Data dependent features and data independent features.  & AdaBoost & 0.56 & 0.86 & 0.86 & 40  \\ \hline
    
\citet{abozinadah2016improved} \citeyear{abozinadah2016improved}& Twitter &  Arabic  & Abusive & Un-Supervised & Lexicon, bag of words (BOW), N-gram & SVM & 0.96 & 0.96 & 0.96 & 13 \\ \hline
    
\citet{abozinadah2017statistical} \citeyear{abozinadah2017statistical} & Twitter & Arabic & Abusive & Supervised & PageRank (PR) algorithm, Semantic Orientation (SO) algorithm, statistical measures. & SVM & 0.96 & 0.96 & 0.96 & 14 \\ \hline
    
\citet{mubarak2017abusive} \citeyear{mubarak2017abusive} & Twitter,  Arabic News Site &  Arabic & Abusive, Offensive & Un-supervised & unigram and bigram, Log Odds Ratio (LOR), Seed Words lists None.  & Just performed extrinsic evaluation & 0.98 & 0.45 & 0.60 & 126\\ \hline

\citet{haidar2017multilingual} \citeyear{haidar2017multilingual}&  Facebook, Twitter   & Arabic & Cyber- bullying (Yes, No) & Supervised & Tweet to SentiStrength, Feature Vector & SVM & 0.93 & 0.94 & 0.92 & 26  \\ \hline

\citet{abdelfatah2017unsupervised} \citeyear{abdelfatah2017unsupervised}& Twitter  & Arabic & Violent (Violent, Nonviolent)  & Un-supervised & Sparse Gaussian process latent variable model, morphological features  Vector Space Model & K-means clustering & 0.56 & 0.60 & 0.58 & 7 \\ \hline

\citet{alfina2017hate}  \citeyear{alfina2017hate} & Twitter & Indonesian & Hate (Hate , Non-hate) & Supervised & BOW and n-gram & Random Forest & - & - & 0.93 & 57\\\hline

\citet{ozel2017detection} \citeyear{ozel2017detection}& Twitter, Instagram & Turkish & Hate & Supervised  & BOW & M-Naïve Bayes & - & - & 0.79 & 23\\\hline
    
\citet{alakrot2018towards} \citeyear{alakrot2018towards}  & YouTube &  Arabic & Offensive, In-offensive & Supervised & N-gram & SVM & 0.88 & 0.80 & 0.82 & 25 \\ \hline
    
\citet{Albadi2018are}  \citet{Albadi2018are}& Twitter &  Arabic & Religious hate, Not hate & Supervised & Word embeddings(AraVec)  & GRU-based RNN & 0.76 & 0.78 & 0.77 & 51  \\ \hline

\citet{alshehri2018think} \citeyear{alshehri2018think}& Twitter  & Arabic & Adult, Regular user & Supervised & Lexicon, N-grams, bag-of-means (BOM) & SVM & 0.70 & 0.93 & 0.78 & 9  \\ \hline

\citet{jaki2019right} \citeyear{jaki2019right}& Twitter & German & Radicalization ( Muslim, Terrorist, Islamo fascistoid) & Un-Supervised & Skip grams and Character trigrams & K-means, single-layer averaged Perceptron & 0.84 & 0.83 & 0.84 & 16\\\hline

\citet{alami2020lisac} \citeyear{alami2020lisac}& Twitter & Arabic & - & Supervised & - & AraBERT & 90 & - & - & -\\\hline

\citet{m1} \citeyear{m1}& Twitter & Tamil-English, Malaylam-English (Code-mix ) & - & Supervised & - & XLM-RoBERTa + mBERT& TENG 90, MENG 77 & - & - & -\\\hline

\citet{polignano2019alberto} \citeyear{polignano2019alberto}& Twitter & Italian & - & Supervised & - & AlBERTO & 90 & 94 & - & -\\\hline

\citet{wang2020galileo} \citeyear{wang2020galileo}& Twitter & English Turkish Arabic Danish Greek & - & Supervised & - & XLM-RoBETa base and lagre & 92 \newline 82 \newline 90 \newline 81 \newline 83 & - & - & -\\\hline

       \end{tabular}}
\end{table}

\clearpage

\subsubsection{Overview of Deep-learning records}
\label{Deep-learning records}
Our systematic review has identified 96 documents related to the application of deep-learning technology/models to the task of automatic hate speech detection. In the sequel, we have analyzed two crucial aspects: the architecture of the deep learning model and the features employed. 

\begin{figure}
    \centering
    \includegraphics[width=.7\linewidth,]{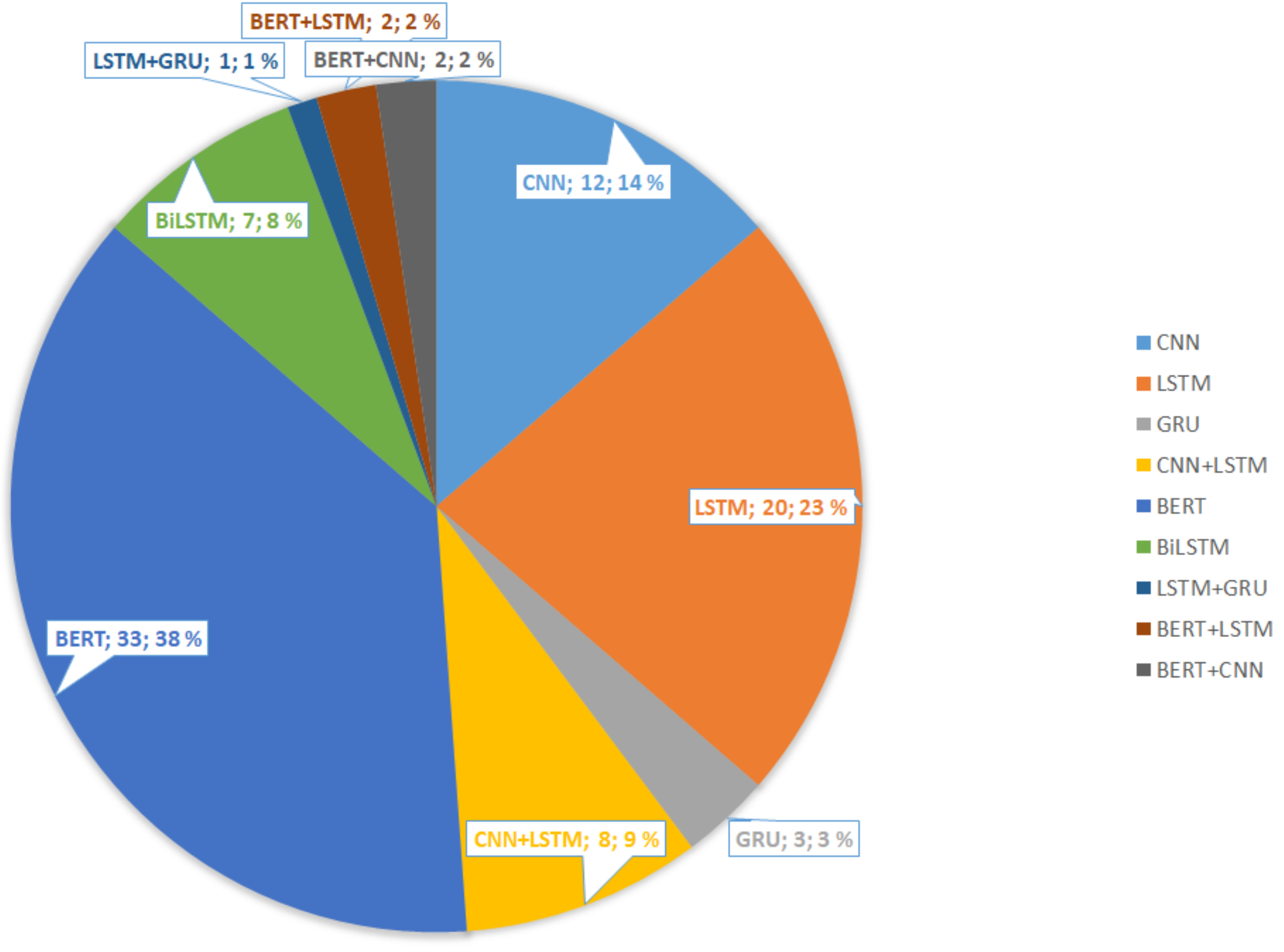}
    \caption{Statistics of previous work based on the percentage of different deep learning algorithms used ( e.g., CNN, LSTM, BERT etc).}
    \label{fig:statistic_deep_al}
\end{figure}

Figure~\ref{fig:statistic_deep_al} summarizes the finding in terms of the percentage of the various deep-learning algorithms employed. Notice that BERT (33\%) becomes prevalent though it was only introduced recently in 2019. The next most popular deep-learning models are LSTM and CNN, which covered 20\% and 12\% of total identified records. Hybrid models (combination of multiple models) depicted in the plot are exemplified by BERT+CNN (2\%), LSTM+CNN(9\%), LSTM+GURU(1\%) and BERT+LSTM(2\%).

In Table~\ref{tab:deeptab}, we tracked different architectures of deep learning employed in the identified records. Surprisingly a total of 24 different deep learning models and feature combinations were found, possibly highlighting the diversity of research's attempts to produce novel architectures building on existing HS detection architectures. Most of the architectures used two steps: (i) word embedding layer employing models such as Word2Vec, FastText, GloVe; (ii) deep learning layer, where one distinguishes, among others,  CNN, LSTM, GRU architectures. 

\begin{table}
    \centering
    \caption{Algorithms and feature used in the papers  related to deep-learning analysis.} 
    \label{tab:deeptab} 
    \scalebox{.85}{\begin{tabular}{|p{5cm}|p{2cm}|p{5cm}|p{2cm} |}
    \hline
    \bf{Architecture name} &  \bf{Frequencies} & \bf{Architecture nam} & \bf{Frequencies}\\ \hline

Word2Vec + LSTM & 6 &  Word2Vec + CNN & 4  \\\hline
RandomEmbedding + LSTM & 4 &  RandomEmbedding + CNN & 3  \\\hline
Word2Vec + BiLSTM & 2 &  Word2Vec + CNN + LSTM & 4  \\\hline
FAstText + LSTM & 4 &  FastText + CNN & 3  \\\hline
FAstText + GRU & 4 &  FastText + GRU & 1  \\\hline
FAstText + GRU & 4 &  FastText + GRU & 1  \\\hline
AraVec + LSTM & 4 &  AraVec + CNN & 3  \\\hline
AraVec +CNN+ LSTM & 1 &  Skip-gram + CNN+LSTM & 1  \\\hline
ELMO + CNN & 1 &  BERT + CNN & 3  \\\hline
ELMO + BERT & 1 &  SKIP-GRAM + CNN & 2  \\\hline
BERT Base & 7 &  BERT Large  & 8  \\\hline
GloVe+CNN & 2 &  GloVe+GBDT+CNN & 1  \\\hline
CNN+CNN+CNN & 2 &  CNN+BiGRU & 1  \\\hline

\end{tabular}}
\end{table}

\begin{figure}
    \centering
    \includegraphics[width=.8\linewidth,]{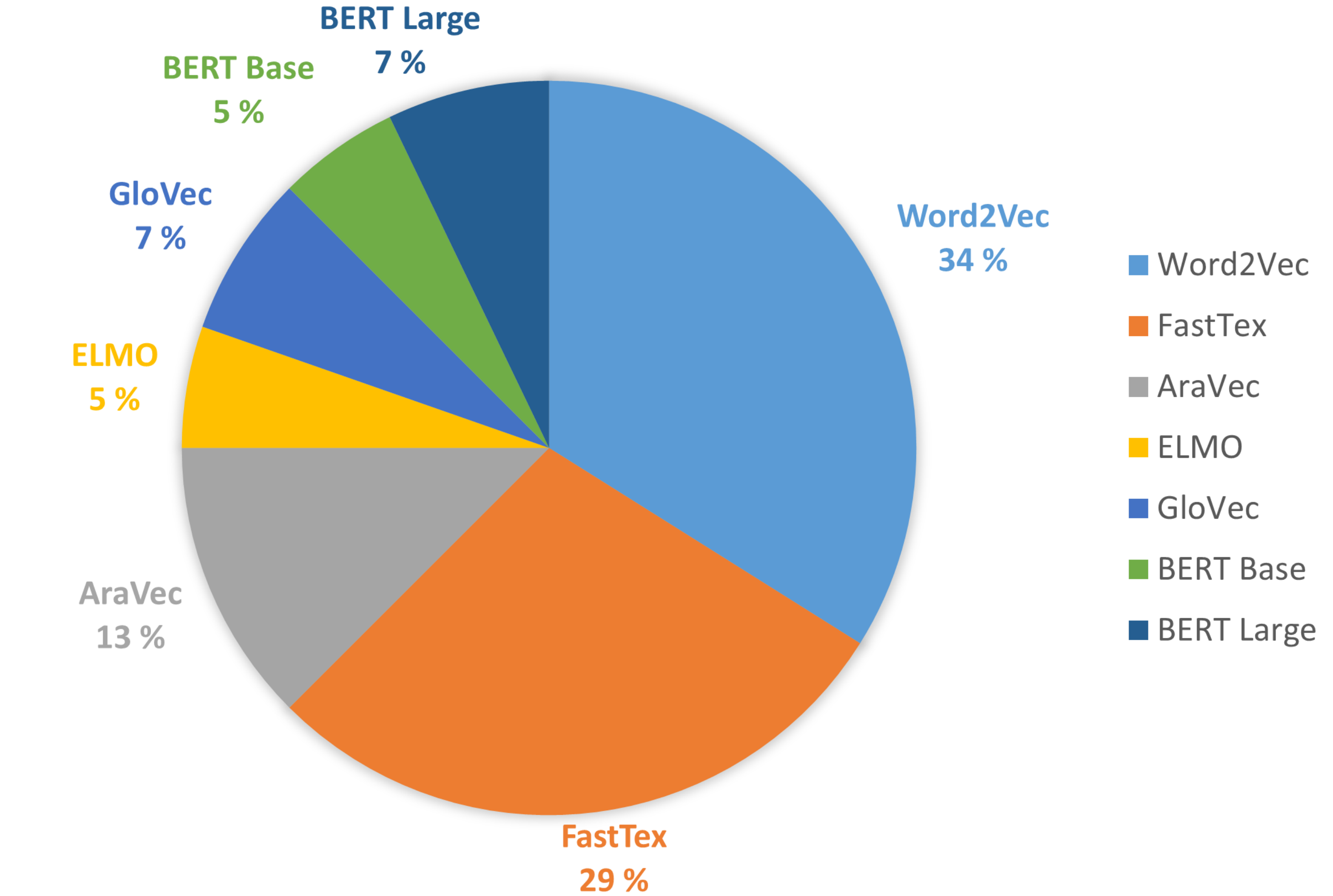}
    \caption{Statistics of word embedding (e.g., Word2Vec, EELMO, FastText, BERT etc ) in Deep learning related records.}
    \label{fig:statistic_deep}
\end{figure}

During the course of this analysis, four significant insights are distinguished: \\
\textbf{I) Comparison between non deep-learning and deep-learning models} revealed that deep-learning models outperformed popular classifiers (NB, LR, RF, SVM) in most studies \cite{dowlagar2021hasocone} \cite{al2021detection} \cite{jahan2020team} \cite{badjatiya2017deep}. An early work by \citeauthor{badjatiya2017deep} \cite{badjatiya2017deep} has compared HS detection with different ML models (LR , SVM, GBDT) showed that deep learning models using either CNN or LSTM model performed on average 13-20\%  better that LR, SVM or GBDT models. Another recent work by \citeauthor{al2021detection} \cite{al2021detection} compared SVM as a baseline model to CNN, CNN+LSTM, GRU, CNN+GRU. The authors found that, in all cases, CNN outperformed the baseline model by at least 7\% in terms of accuracy. 
\\

\textbf{II) Comparison between CNN, LSTM, BiLSTM, and GRU models} revealed mixed results in terms of which deep-learning architecture performs most. For instance, a comparison between CNN and LSTM architecture showed a better CNN performance results in \cite{jahan2020team}, whereas \citeauthor{badjatiya2017deep} \cite{badjatiya2017deep} found that LSTM architecture performs better than CNN. Besides, the difficulty in data full reconstruction, the difficulty to reproduce exact preprocessing stages or use of distinct embedding can render such comparison more challenging. Yin et al. \cite{yin2017comparative} conducted a comparative study between two deep-neural network architectures: “RNN (with GRU and LSTM layers) and CNN”. They found that RNN is more suited for the long-ranged context dependencies, while CNN sounds better in extracting local features. They also revealed that GRU performs better in case of long sentences. 
However, several other papers suggested that the concatenation of two or more deep learning models perform better than a single deep learning model \cite{al2021detection} \cite{zhou2020deep}  \cite{kapil2020deep} \cite{shruthi2020hate} \cite{kumar2021comment}. For example, CNN+LSTM and CNN+GRU both performed better than the single application of LSTM and CNN \cite{al2021detection}. \citeauthor{zhou2020deep} \cite{zhou2020deep} suggested a fusion of three CNN models with different parameters as a viable way to improve the performance of hate speech detection. \citeauthor{kapil2020deep} \cite{kapil2020deep} compared CNN, LSTM, and LSTM+GRU architectures for HS detection in five commonly used datasets, and concluded that in all cases, LSTM+GRU outperform a single LSTM and CNN model by 2-3\%. Similarly, \citeauthor{shruthi2020hate} \cite{shruthi2020hate} proposed a hybrid fusion architecture BiLSTM + Random Embedding + TF-IDF + LR that achieves at least a 12\%  increase in performance compared to a single architecture. 

\textbf{III) Comparison between word embedding} revealed a lack of comparative studies in this area as well. However, from Table~\ref{tab:deeptab}, we can see that Word2vec and FastText were regularly used with different deep learning architectures. Although this popularity does not entail systematically a better performance score. For instance, \citeauthor{saleh2020detecting} \cite{saleh2020detecting} compared word2Vec, GloVe, and Google NewsVec, and showed that word2Vec performed only 1\% better than others. However, when compared to BERT model, BERT-Large showed the best accuracy and F1 score. Another work by \citeauthor{rizos2019augment} \cite{rizos2019augment} comparing FastText, GloVe and GoogleNewsVec revealed no significance accuracy among these embeddings.

On the other hand, since Word2Vec, FastText, GloVe use a vector representation to represent words in a way that captures semantic or meaning-related relationships as well as syntactic or grammar-based relationships, this also bears inherent limitation in the sense that this cannot capture polysemy relationship. That is, for the same word, even if it has different meanings in different contexts, the corresponding vector representation is unchanged. We shall mention the merits of the recently introduced ELMO word embedding model, which was designed to overcome the aforementioned shortcoming. A recent work by \citeauthor{zhou2020deep} \cite{zhou2020deep} using ELMO embedding showed a better performance compared to CNN. However, since ELMO is a relatively new,  the in-depth comparison with other embedding model is still in its infancy. This leaves the door wide for future experiments
\\

\textbf{IV) The rise of BERT} is a striking trend that can be seen in Figure~\ref{fig:statistic_deep}, which testifies of its  popularity in hate speech detection community (38\% share of deep-learning models) in the past five years. This demonstrates the importance of this model as a key state-of-the-art method in the field. Several works explored BERT performance in HS detection \cite{nikolov2019nikolov} \cite{ranasinghe2019brums} \cite{saleh2020detecting} \cite{dowlagar2021hasocone} where almost all authors who compared BERT model to other deep learning models concluded on the superiority of BERT architecture. For instance, \citeauthor{ranasinghe2019brums} \cite{ranasinghe2019brums} compared BERT to FastText, CNN, and LSTM. Similarly, \citeauthor{saleh2020detecting} \cite{saleh2020detecting} compared BERT to BiLSTM and LSTM where BERT showed a significance performance increase. Also, BERT model achieved top performance in the multilingual tasks in \cite{dowlagar2021hasocone} \cite{polignano2019alberto} \cite{alami2020lisac} \cite{m1} \cite{wang2020galileo}. 

Some research performed a comparison among different BERT models. For instance, \citeauthor{wang2020galileo} \cite{wang2020galileo} found that BERT-large model outperformed BERT-base model for HS detection. In addition, two different BERT architecture concatenations (e.g., XLMR-B + mBERT) are found to perform better than a single pretrained BERT\cite{m1}. Furthermore, BERT architecture can be trained in a such a way that it can perform a specific task. For example, \citeauthor{gonzalez2021twilbert} \cite{gonzalez2021twilbert} proposed TWiLBERT, a specialization of BERT architecture for the Spanish and Twitter domain. They performed an extensive evaluation of TWilBERT models on 14 different text classification tasks, such as irony detection, sentiment analysis and emotion detection. The results obtained by TWilBERT outperformed the state-of-the-art systems and mBERT. Another work by \citeauthor{caselli2020hatebert} \cite{caselli2020hatebert} introduced HateBERT, a re-trained BERT model for abusive language detection in English. The model was trained on a large-scale dataset of Reddit comments in English from communities banned for being offensive, abusive, or hateful. In all cases, HateBERT outperformed the corresponding general BERT model. 

Besides, some other language-specific BERTs models developed over time for monolingual outperformed multilingual model mBERT: AraBERT (Arabic) \cite{antoun2020arabert}, AlBERTo (Italian) \cite{alberto}, FinBERT (Finnish) \cite{finbert}, CamemBERT(French) \cite{camembert}, Flaubert (French \cite{flaubert}), BERT-CRF (Portuguese) \cite{bert-crf}, BERTje (Dutch) \cite{bertje}, RuBERT (Russian) \cite{rubert} and  BERTtweet (A pre-trained language model for English Tweets) \cite{bertweet}. However, to best of our knowledge, not every model has yet been tested for HS domain except AraBERT \cite{alami2020lisac} \cite{djandji2020multi} and AlBERTo \cite{polignano2019alberto} which shown better performance for HS detection. 

\begin{table}
    \centering
    \caption{Summary of some key contributions in HS detection by using Deep learning method and their respective
results, in the metrics: Precision (P), Recall (R), F1-Score (F), Citation (C).} 
    \label{tab:Arabicdata} 
    \scalebox{.75}{\begin{tabular}{|p{2.5cm}|p{2.3cm}|p{2.2cm}|p{2cm} |p{4cm}|p{2.3cm}|l|l|l|l|}
    \hline
    \bf{Author, Year} &  \bf{Platform} & \bf{Type} & \bf{ML Approach}& \bf{Features Representation} & \bf{Algorithm}  & \bf{P}& \bf{R} & \bf{F}& \bf{C}  \\ \hline

 \citet{badjatiya2017deep} \citeyear{badjatiya2017deep} &  Twitter, 16k & Hate speech & Supervised & FastText, Random embedding, GloVe & CNN, LSTM, GBDT & .93 & .93 & .93 & 503 \\\hline
 
 \citet{yin2017comparative} \citeyear{yin2017comparative} &  - & - & Supervised & & CNN, GRU and LSTM & .94 & - & - & 534 \\\hline

\citet{rizos2019augment} \citeyear{rizos2019augment}&  Twitter, 24k & Hate speech & Supervised & FastText, Word2Vec, GloVe & CNN, LSTM, GRU & - & - & .69 & 20 \\\hline

\citet{kamble2018hate} \citeyear{kamble2018hate}& Twitter, 3.8k & Hate speech & Supervised &  Word2Vec & LSTM, BiLSTM, CNN & .83 & .78 & .80 & 14 \\\hline

\citet{ranasinghe2019brums} \citeyear{ranasinghe2019brums}& Twitter, & Hate speech & Supervised & FastText & LSTM, GRU, BERT & - & - & .78 & 16 \\\hline

\citet{faris2020hate} \citeyear{faris2020hate} & Twitter, (Arabic) & Hate speech & Supervised & Word2Vec, Aravec & CNN+LSTM & .65 & .79 & .71 & 3 \\\hline

\citet{al2021detection} \citeyear{al2021detection}, Springer & Twitter, 11k (Arabic) & Hate, Racism, Sexism & Supervised & Keras word embeding & LSTM,GURU, CNN+GRU, CNN+LSTM & .72 & .75 & .73 & 0 \\\hline

\citet{duwairi2021deep} \citeyear{duwairi2021deep}, Springer &  Twitter, 9k,2k (Arabic) & Hate, Hate, Abusive,Misogyny,
Racism, Religious Discrimination, & Supervised & SG, CNN, CBOW & CNN, CNN-LSTM, and BiLSTM-CNN & .74 & -& - & 0 \\\hline

\citet{dowlagar2021hasocone} \citeyear{dowlagar2021hasocone}, arXiv & Twitter,(English, German, Hindi) & Hate, Offensive & Supervised & ELMO & BERT, Multilingual-BERT & .83 & -& 83 & 2 \\\hline
   
\citet{sigurbergsson2019offensive} \citeyear{sigurbergsson2019offensive} & Twitter, Reddit, newspaper comments (Danish) & Hate, Offensive, Not, Target,Individual, Group & Supervised & - & AUX-Fast-BiLSTM  & - & - & .67& 53\\\hline

 \citet{ousidhoum2019multilingual} \citeyear{ousidhoum2019multilingual} &  Twitter & Sexual orientation, Religion, Disability, Target Group & Supervised & BOW & LR, biLSTM   & - & - & 80 & 35\\\hline

\citet{mulki2019hsab} \citeyear{mulki2019hsab}&  Twitter & Abusive, Hate & Supervised & - & CNN and BiLSTM-CNN  & .74 & - & -& 29\\\hline

\end{tabular}}
\end{table}

Table~\ref{tab:competition} summarizes selected key works from three major recent HS detection competitions SemEval-2019 \cite{zampieri2019semeval}, SemEval-2020 \cite{zampieri2020semeval}, and Hasoc-2020 \cite{mandl2020overview}.

In SemEval-2019, Task A (offensive language detection) was the most popular sub-task with 104 participating teams. Among the top-10 teams, seven used BERT with variations in the parameters and in the pre-processing steps. The top-performing team \citeauthor{liu2019nuli} \cite{liu2019nuli} used BERT-base-uncased with default-parameters, but with a max sentence length of 64 and trained for 2 epochs and achieved 82.9\% F1 score which was 1.4 points better than \citeauthor{nikolov2019nikolov} \cite{nikolov2019nikolov}. Although, the difference between the next five systems, ranked 2-6, is very marginal (less than one percent (81.5\%-80.6\%)). The top nonBERT model by \citeauthor{mahata2019midas} \cite{mahata2019midas} was ranked fifth. They used an ensemble of CNN and BLSTM+BGRU, together with Twitter word2vec embeddings and token/hashtag normalization.

In semEval-20, 145 teams submitted official runs on the test data and 70 teams submitted system description papers. The best team \citeauthor{wiedemann2020uhh} \cite{wiedemann2020uhh} achieved an F1 score of 0.9204 using an ensemble of ALBERT models of different sizes. The second team \citeauthor{wang2020galileo} \cite{wang2020galileo} achieved an F1 score of 0.9204, and it used RoBERTa-large that was fine-tuned with the dataset in an unsupervised way. The third team \citeauthor{dadu2020team} \cite{dadu2020team}, achieved an F1 score of 0.9198, using an ensemble that combined XLM-RoBERTa-base and XLM-RoBERTa-large trained on Subtask A data for all languages. The top-10 teams were close to each other and employed BERT, RoBERTa or XLM-RoBERTa models; sometimes CNNs and LSTMs were also been mentioned either for comparison or hybridization purpose.

\begin{table}
    \centering
    \caption{Best architecture from previous competition and their respective
results, in the metrics: Precision (P), Recall (R), F1-Score (F), Citation (C).} 
    \label{tab:competition} 
    \scalebox{.75}{\begin{tabular}{|p{3cm}|p{3cm}|p{2.6cm}|p{2cm} |p{2.3cm}|p{3cm}|l|l|}
    \hline
    \bf{Organiser}& \bf{Author, Year} &  \bf{Platform} &\bf{Language}& \bf{Features Representation} & \bf{Algorithm}  &  \bf{F1 (\%)}& \bf{C}  \\ \hline
    
SemEval-19 Task 6  & \citet{liu2019nuli}, 2019 & Twitter 14k & English & - & LSTM, BERT & 82.9 & 70 \\\hline

SemEval-19 Task 6  &  \citet{nikolov2019nikolov}, 2019  & Twitter 14k & English & - & NB, CNN, LR, SVM, BERT-Large   & 81.5 & 35 \\\hline
SemEval-19 Task 6  &  \citet{mahata2019midas}, 2019&  Twitter 14k & English & Word2vec  &  CNN, BLSTM+BGRU   & 80.6 & 70 \\\hline

SemEval-20 Task 12  &  \citet{wiedemann2020uhh}, 2020&  Twitter, 14k & English & - & BERT-base, BERT large, RoBERTa, XLM-RoBERTa, ALBERT & 92 & 5 \\\hline
SemEval-20 Task 12  &  \citet{wang2020galileo}, 2020&  Twitter, 14k & English & - & XLM-RoBERTa base, XLM-RoBERTa large  & 91.9 & 1 \\\hline
SemEval-20 Task 12  &  \citet{dadu2020team}, 2020&  Twitter, 14k & English & - & XLM-RoBERTa  & 91.8 & 1 \\\hline

HASOC2020  &  \citet{mishraa2020iiit_dwd}, 2020 & Twitter & English & GloVe &  LSTM  & 51 & 0 \\\hline
HASOC2020  & \citet{kumar2020comma}, 2020& Twitter & German & - &  BERT, DistilBERT and RoBERTa  & 52 & 0 \\\hline

HASOC2020  & \citet{raja2021nsit}, 2020& Twitter & Hindi & FasText & BiLSTM, CNN & 52 & 0 \\\hline
\end{tabular}}
\end{table}

Over 40 research groups participated in Hasoc-2020 competition. The top ranked submission for Hindi-hate speech detection, used a CNN with FastText embeddings as input \cite{raja2021nsit}. The best performance for German hate speech detection task was achieved using a fine-tuned versions of BERT, DistilBERT and RoBERTa \cite{kumar2020comma}. Similarly, the top performance in English-language HS detection was based on a LSTM architecture with GloVe embeddings as input \cite{mishraa2020iiit_dwd}.


\section{Resources for hate speech detection }
\label{resources}
In the conducted literature review, several useful resources were identified. In this section, we represent the datasets and open source projects.

\subsection{Hate speech available datasets}
Regarding the datasets, we found 69 datasets in 21 different languages. In this section, we summarize the most used dataset attributes and statistics in Tables \ref{tab:Englishdatasets} and \ref{tab:multilingualdatasets}. This includes dataset names (some names are based on papers title), publication year, dataset source link \footnote{All datasets' link last access was on 03-march-2021}, dataset sizes, the ratio of offensive contents, the class used for annotation, and datasets' language.  We noticed that many authors collected their datasets from social media and then annotated them manually based on task requirements. Several annotations have been carried out with experts \cite{waseem2016hateful} \cite{zhang2019hate} \cite{kumar2018aggression}, native speakers \cite{gao2017detecting}, volunteer \cite{warner2012detecting}, or through crowd-sourcing from anonymous users \cite{davidson2017automated}\cite{wulczyn2017ex}\cite{zampieri2019semeval}. Below we present the primary findings of this analysis.

\begin{figure}
    \centering
    \includegraphics[width=.95\linewidth,]{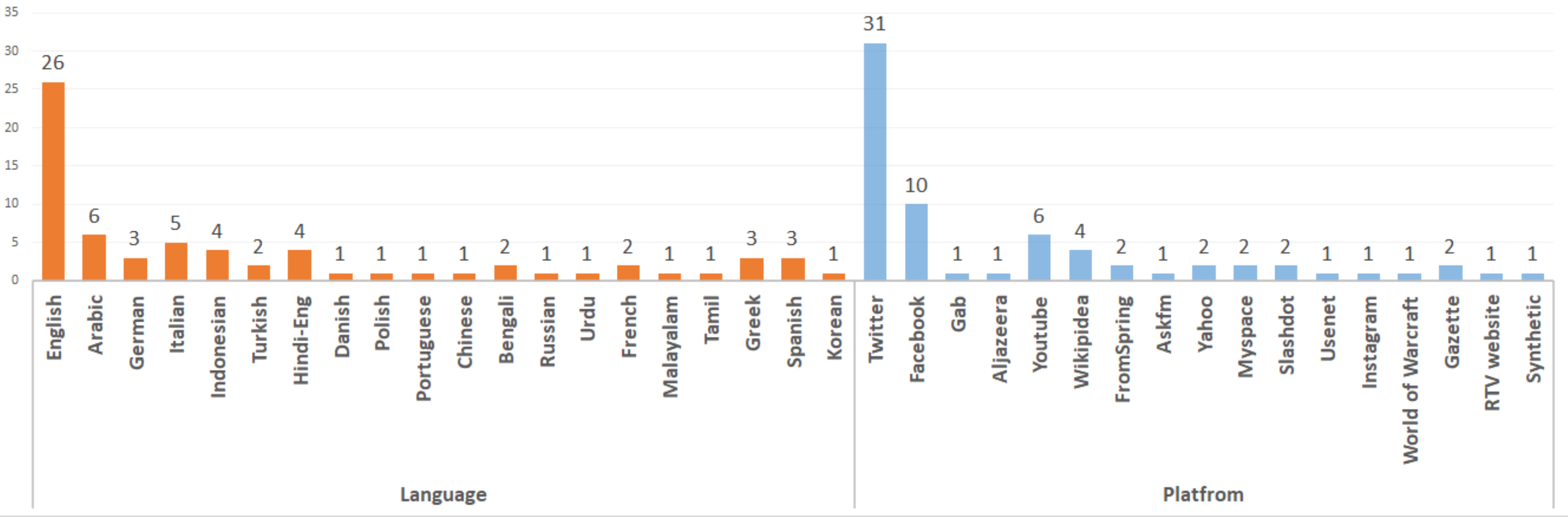}
    \caption{Number of Datasets from different Language and Social Media.}
    \label{fig:languageand}
\end{figure}

I) \textbf{Datasets language and platform:} Among 69 datasets of 21 different languages, see Figure \ref{fig:languageand}], English dominates by far others, representing 26 datasets alone. However, Arabic, German, Hind-English, Indonesian and Italian are represented in a total of 6, 3, 4, 4 and 5 open datasets, respectively. The rest of the languages have low presence in this set of open dataset. All datasets were collected from different social media platforms (Twitter, Facebook, Youtube, etc.), with exception to \citet{chung2019conan}'s dataset where some portion were synthetically produced. Twitter is shown to be the most popular platform for collecting hate speech datasets (45\% of total datasets were collected from Twitter).  Facebook is the second most popular source. The rest of the SM has only been used few times. 
 
An interesting code-mixed dataset by  \citeauthor{mathur2018did} would be an example for targeting users who are active in SM and use the mixed form of language. This dataset has highlighted the predominance of Hindi–English code–mixed data representing the large spread of mixed forms and Hindi words written in Latin script in a non-formal online communication among Indians SM users. Similar code-mixed dataset work is also done in Tamil-English and Malayalam-English using BERT, which achieved a 90\% F1 score \cite{m1}.

(II) \textbf{Datasets sources:} Most of the dataset source repositories are available on GitHub. Therefore, nearly all datasets were publicly available. However, those dataset collected from Twitter have only Twitter Id instances which should be used to retrieve the full tweet messages. Since many tweets might be deleted over time, one may expect that the reconstruction of the full dataset may not be possible. 

\begin{figure}
    \centering
    \includegraphics[width=\linewidth,]{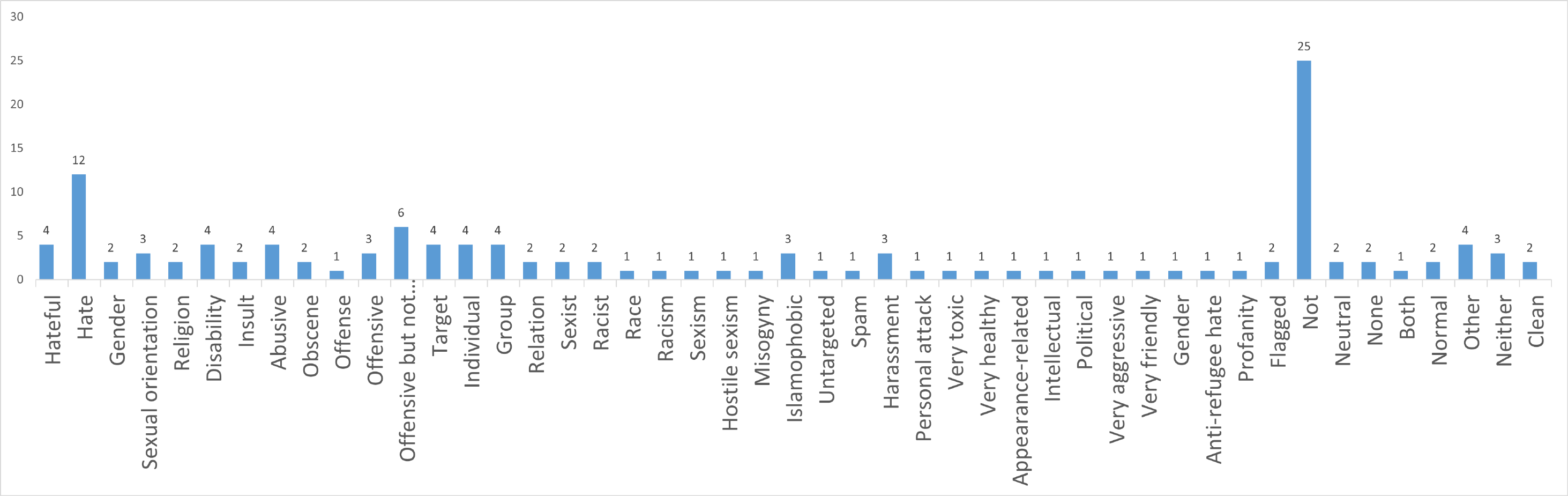}
    \caption{Number of Classes used for annotation.}
    \label{fig:class}
\end{figure}

(III) \textbf{Annotation classes:} Figure \ref{fig:class} illustrates the diversity of annotations in the original datasets (e.g., hate, offensive, race, gender, sexism, misogyny, toxicity,  group. target, political, etc.).  This diversity of annotations translates the variety of hate speech categories and the academic willingness to explore this rich panorama. However, from Tab.~\ref{tab:Englishdatasets} and \ref{tab:multilingualdatasets}, one notices that most of these annotations were based on binary classification (e.g., hate versus non-hate, racism versus non-racism, etc.) \cite{Albadi2018are} \cite{alakrot2018dataset}\cite{sanguinetti2018italian} \cite{bohra2018dataset}\cite{ribeiro2018characterizing}\cite{qian2019benchmark}\cite{elsherief2018peer} \cite{bretschneider2016detecting}\cite{rosa2018deeper}\cite{gao2017detecting}. Ternary class levels as well (e.g., Hate, Abusive, Normal) are explored in \cite{mulki2019hsab} \cite{mubarak2017abusive}\cite{waseem2016hateful}\cite{wulczyn2017ex}\cite{ousidhoum2019multilingual}. Some authors used a larger number of classes and sub-classes (up to six) as in  \cite{ousidhoum-etal-multilingual-hate-speech-2019}\cite{zampieri2019semeval}\cite{mandl2019overview}\cite{rezvan2018quality}\cite{bretschneider2016detecting}.
In summary, one can distinguish three strategies of annotation scheme. The first one is a binary scheme: two mutually exclusive events (typically yes/no) to mark the presence or absence of  HS (or a category of HS). 

The second one advocates a non-binary scheme with a fixed number of mutually exclusive classes, accounting either for different shades of a given HS category, such as strong hate, weak hate, no hate \cite{de2017offensive}, overtly aggressive, covertly aggressive, not aggressive \cite{kumar2018aggression}, or for several classes at the same time, such as racism, sexism, racism and sexism, none \cite{waseem2016hateful}.

The third strategy features multi-level annotation, with finer-grained schemes accounting, for instance, for the type of hate speech, its severity, and the target group. This is the most complex annotation scheme and typically involves several different traits and a scale of variation. For example, \citeauthor{gomez2020exploring} distinguish between racist, sexist, homophobic, religion-based attack, as well as the community targeted by the attack in the annotation process. \citeauthor{nobata2016abusive} \cite{nobata2016abusive} discriminate between clean and abusive language, where abusive is labeled as hate speech, derogatory or profane. Basile et al. \cite{basile2019semeval} adopt a three-layer binary annotation for HS, aggressiveness, and nature of the target (individual or group).

(IV) \textbf{Dataset size and ratio of abusive contents:}

\begin{figure}
    \centering
    \includegraphics[width=.5\linewidth,]{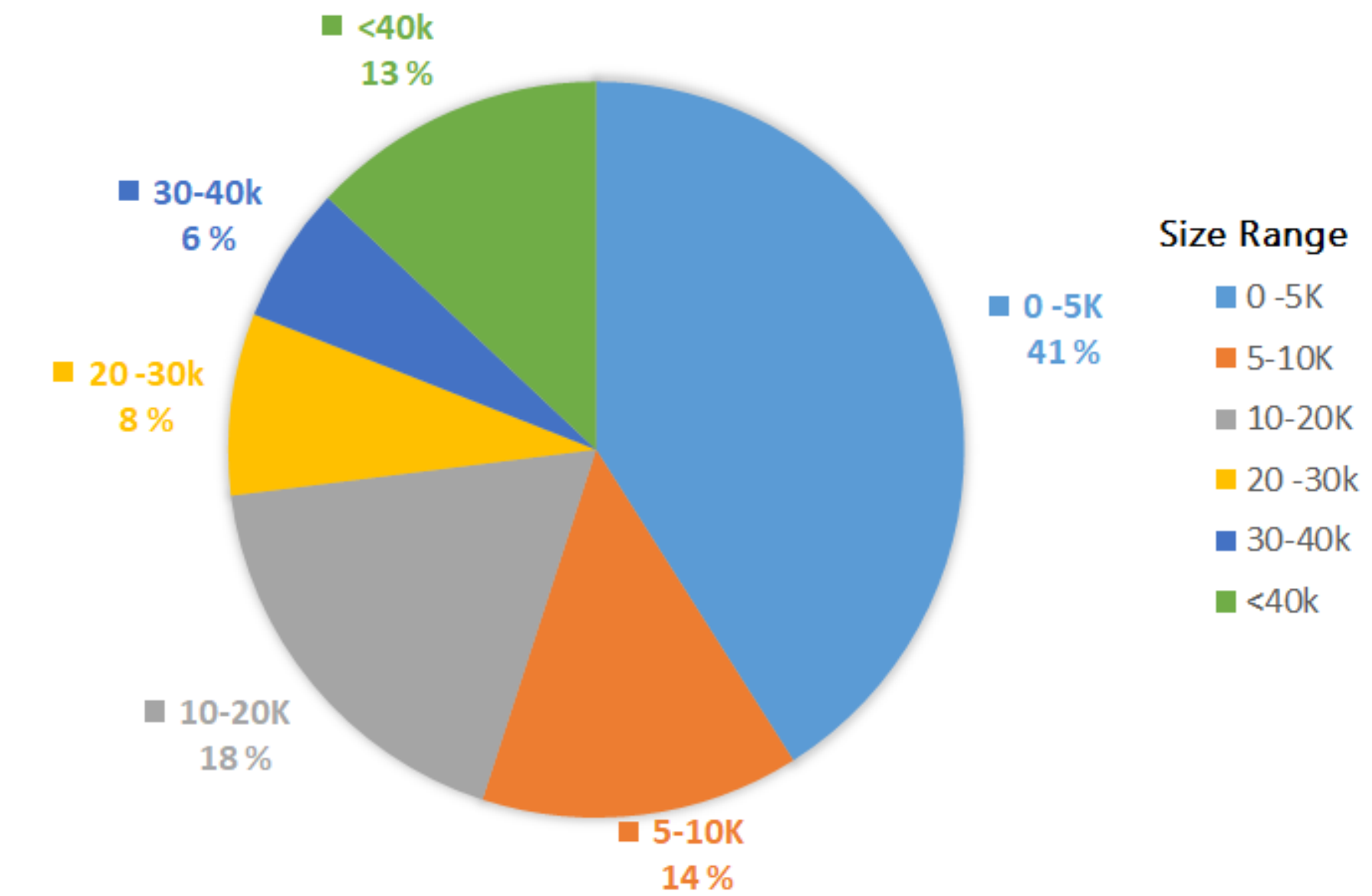}
    \caption{Statistics of dataset size. k represent thousand.}
    \label{fig:dataset_size_ratio}
\end{figure}

\begin{figure}
    \centering
    \includegraphics[width=.5\linewidth,]{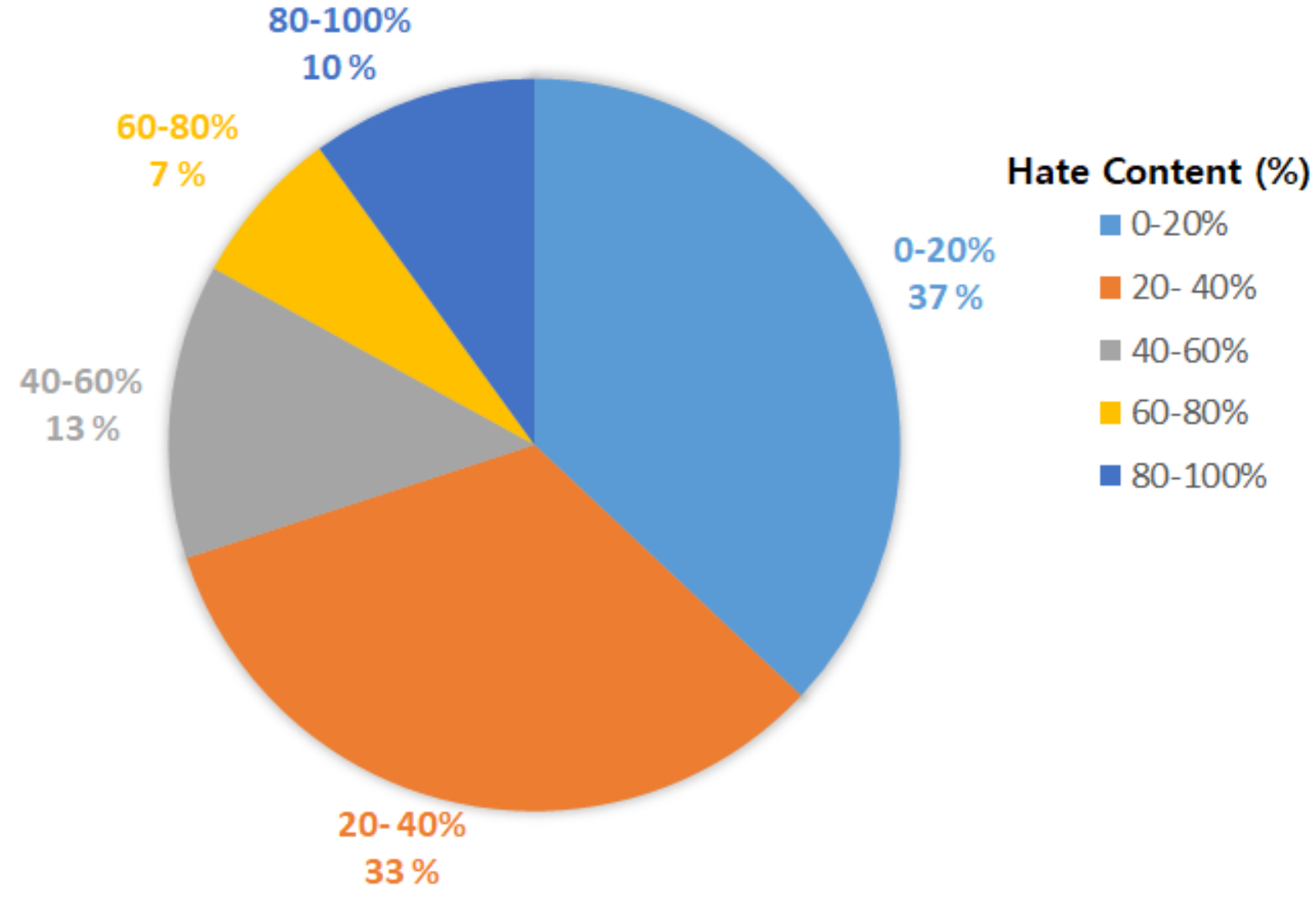}
    \caption{Statistics of the proportion of abusive content contained in hate speech dataset.}
    \label{fig:dataset_ratio}
\end{figure}

Figure \ref{fig:dataset_size_ratio} and \ref{fig:dataset_ratio} show the statistic of dataset size and the ratio of offensive content, respectively. We can see 41\%  of datasets are relatively of a small size (only (0-5)k posts). 14\% have (5-10)k sentences. Therefore, most of the dataset (55\%) can be cast into a very small size category, indicating the challenges behind acquiring large-scale labeled data for hate speech detection purpose. On the other hand, we have not found any balance dataset. This can lead to overfitting and harm generalisability, especially for deep learning models \cite{goodfellow2016deep}.

Another critical factor that may affect the training process of the model is the ratio of classes. In this regards, we noticed a high number (37\%) of the datasets contain significantly less than 20\% of offensive content. However, the rest of the datasets ( 63\%) have more than 20\% offensive content, which could be considered as decent for training purposes. 

\begin{figure}
    \centering
    \includegraphics[width=1\linewidth,]{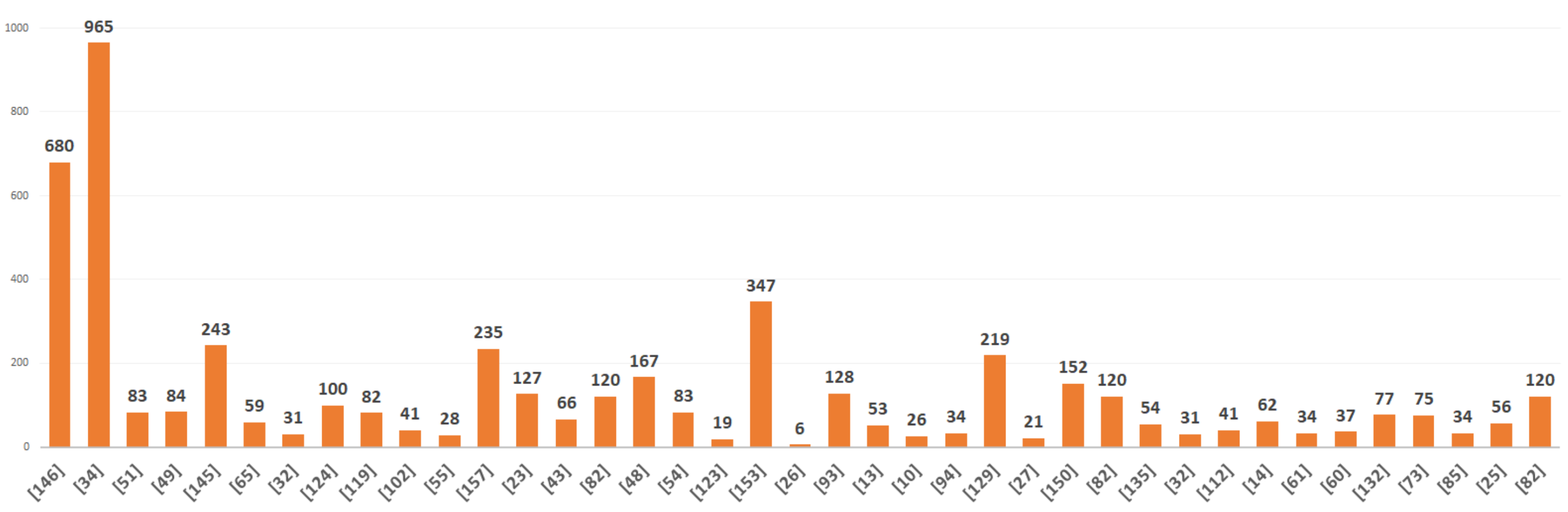}
    \caption{Number of citations of reference papers of dataset.}
    \label{fig:citation_dataset}
\end{figure}

(V) \textbf{Number of Citations\footnote{All citations last updated on 03-Mar-2021}:} 
We collected the number of citations for each dataset source document in Google Scholar and concluded that most dataset sources were cited more than 50 times (Figure~\ref{fig:citation_dataset}). \citet{davidson2017automated} is the most cited papers that have presented a dataset of 24802 tweets, which are manually annotated by CrowdFlower (CF) workers. Workers were asked to label each tweet as one of three categories: hate speech, offensive but not hate-speech, or neither offensive nor hate speech. Three or more people participated in each tweet annotation process. They have used the majority decision for each tweet to assign a label. This dataset has only 6\% hate speech, 76\% offensive but non-hate, and the rest of the tweets were neutral.

The second most cited dataset created by \citet{waseem2016hateful} contains 16,914 tweets where 3,383 are related to sexist, 1,972  to racist, and  11,559 to neither sexist nor racist. The authors manually annotated their dataset, after which it was sent to an outside annotator (a 25 year old woman studying gender studies and a nonactivist feminist) to review their annotation.
\\

The above two popular datasets provide some insights into the dataset annotation process and criteria. We also noticed that the citation index does not reflect the quality of the dataset itself, but rather its ease and simplicity, which motivated other researchers to test such dataset in their proposals. There could be many possible factors that may represent the quality of a dataset (e.g., annotation criteria, label definitions, understanding perception, dataset size, the ratio of classes). However, we do not have sufficient experimental evidence to compare these dataset quality indices across various HS domains. To the best of our knowledge, only one study found in our search attempted to estimate the quality metrics and the similarity between datasets. In this respect, Fortuna et al. \cite{fortuna2020toxic} compared the categories across the annotated dataset with respect to both similarity to other categories and homogeneity. For this purpose, average FastText embedding pretrained on Wikipedia was used to represent each category, while intra-dataset class homogeneity index has been put forward to assess category homogeneity. One of their observations is that Davidson's \cite{davidson2017automated} "hate speech" is very different from Waseem's \cite{waseem2016hateful} "hate speech", "racism", "sexism", while being relatively close to Basile's HS dataset \cite{basile2019semeval}.  One of the main conclusions of their experiments is that many different definitions are being used for equivalent concepts, which makes most of the publicly available datasets incompatible.

\begin{table}
    \centering
    \caption{Datasets  for Hate Speech Detection for  English.} 
    \label{tab:Englishdatasets} 
    \scalebox{.78}{\begin{tabular}{|p{4cm}|p{.8cm}|p{1.5cm}|p{1.5cm} |p{1cm}|l|p{5cm}|p{3cm}|}
    \hline
    \bf{Name} &  \bf{Year} & \bf{Link} & \bf{Platform} & \bf{Size}  & \bf{Ratio} & \bf{Class}  & \bf{Ref., Citation} \\ \hline

 
Hate Speech and Offensive Language & 2017 & \href{https://github.com/t-davidson/hate-speech-and-offensive-language} {GitHub} &  Twitter &  24,802 & .06 &  Binary (hate speech, offensive but not hate speech, or neither offensive nor hate speech)
 &  \citet{davidson2017automated}, 965 \\\hline
 
Hate Speech Dataset  & 2018 & \href{https://github.com/Vicomtech/hate-speech-dataset} {GitHub} &  Stormfront &  9,916 & .11 &  Hate, Relation, Not  &  \citet{de2018hate}, 83\\\hline

Predictive Features for Hate Speech Detection & 2016 & \href{https://github.com/ZeerakW/hatespeech} {GitHub} &  Twitter &  16,914 & .32 &  Sexist, Racist, Not &  \citet{waseem2016hateful}, 680 \\\hline

Hate Speech Detection Fox news comments & 2017 & \href{https://github.com/sjtuprog/fox-news-comments} {GitHub} &  Fox News&  1528  & 0.28 &  Binary (Hate / not) &  \citet{gao2017detecting}, 84 \\\hline

Hate Speech Twitter annotations & 2016 & \href{https://github.com/ZeerakW/hatespeech} {GitHub} &  Twitter &  4,033  & 0.16 &  Racism, Sexism  &  \citet{waseem2016you}, 243 \\\hline

Sexism using twitter data & 2016 & \href{https://github.com/AkshitaJha/NLP_CSS_2017} {GitHub} &  Twitter &   712  & 1 &  Sexism   &  \citet{jha2017does}, 59 \\\hline

Misogyny Identification at IberEval 2018 & 2016 & \href{https://amiibereval2018.wordpress.com/important-dates/data/} {GitHub} &  Twitter &   3,977  & 0.47 &  Sexism   &  \citet{jha2017does}, 59 \\\hline

CONAN Multilingual Dataset  of Hate Speech & 2019 & \href{https://github.com/marcoguerini/CONAN} {GitHub} &  Synthetic, Facebook &  1,288  & 1 &  Islamophobia  &  \citet{chung2019conan}, 31 \\\hline

Characterizing and Detecting Hateful Users & 2018 & \href{https://github.com/manoelhortaribeiro/HatefulUsersTwitter} {GitHub} &  Twitter &  4,972  & 0.11 &  Hate, Not hate  &  \citet{ribeiro2018characterizing}, 100 \\\hline

Online Hate Speech (Gab) & 2019 & \href{https://github.com/jing-qian/A-Benchmark-Dataset-for-Learning-to-Intervene-in-Online-Hate-Speech} {GitHub} &  GAB &  33,776 & 0.43 &  Hate, Not hate  &  \citet{qian2019benchmark}, 82 \\\hline

Online Hate Speech (Reddit) & 2019 & \href{https://github.com/jing-qian/A-Benchmark-Dataset-for-Learning-to-Intervene-in-Online-Hate-Speech} {GitHub} &  Reddit &  22,324 & 0.24 &  Hate, Not hate   &  \citet{qian2019benchmark}, 82 \\\hline

Multilingual and Multi-Aspect Hate Speech & 2019 & \href{https://github.com/HKUST-KnowComp/MLMA_hate_speech} {GitHub} &  Twitter &  5,647 & 0.76 &  Gender, Sexual orientation, Religion, Disability   &  \citet{ousidhoum2019multilingual}, 41 \\\hline

HS Detection in Multimodal Publications & 2020 & \href{https://gombru.github.io/2019/10/09/MMHS/} {GitHub} &  Twitter &  149,823 & 0.25 & No attacks to any community, Racist, Sexism, Homophobia, Religion-based attack, Attack to other community   &  \citet{gomez2020exploring}, 28 \\\hline

SemEval-2019 Task 6 & 2019 & \href{https://competitions.codalab.org/\%20competitions/20011} {Link} &  Twitter &  14,100 & 0.33 & Offensive, Not Offensive, Target, Not Target, Individual, Group, Othe   &  \citet{zampieri2019predicting}, 235 \\\hline

Multilingual Detection of HS Against Immigrants and Women & 2019 & \href{https://competitions.codalab.org/competitions/19935} {Link} &  Twitter &  13,000 & 0.4 & Hate, Not Hate, Group, Individual, Aggression, Not Aggression   &  \citet{basile2014enhanced}, 127 \\\hline

Peer to Peer Hate & 2019 & \href{https://github.com/mayelsherif/hate_speech_icwsm18} {GitHub} &  Twitter &   27,330 & 0.98  & Hate, Not Hate &  \citet{elsherief2018peer}, 66 \\\hline

HASOC-2019 (English) & 2019 & \href{https://hasocfire.github.io/hasoc/2019/dataset.html} {GitHub} &  Twitter, Facebook&   7,005 & 0.36  & Hate, Offensive , Neither,  Profane, Targeted, Not Targeted  &  \citet{mandl2019overview}, 120 \\\hline

Twitter Abusive Behavior & 2018 & \href{https://dataverse.mpi-sws.org/dataset.xhtml?persistentId=doi:10.5072/FK2/ZDTEMN} {Link} &  Twitter&  80,000 & 0.18  & Abusive, Hateful, Normal, Spam  &  \citet{founta2018large}, 167 \\\hline

Online Harassment & 2017 & Not available &  Twitter&  35,000 & 0.16  & Harassment, Not Harassment  &  \citet{golbeck2017large}, 83 \\\hline

Personal Attacks & 2017 & \href{https://github.com/ewulczyn/wiki-detox} {GitHub} & Wikipedia &  115,737 & 0.12  & Personal attack, Not Personal attack  &  \citet{wulczyn2017ex}, 347 \\\hline

Toxicity & 2017 & \href{https://github.com/ewulczyn/wiki-detox} {GitHub} & Wikipedia &  100,000 & NA  & very toxic, neutral, very healthy  &  \citet{wulczyn2017ex}, 347 \\\hline

Cyberbullying (World of Warcraft) & 2016 & \href{http://ub-web.de/research/} {GitHub} & World of Warcraft & 16,975 & .01  & Harassment, Not Harassment  &  \citet{bretschneider2016detecting}, 6 \\\hline

Cyberbullying (League of Legends) & 2016 & \href{http://ub-web.de/research/} {GitHub} & League of Legends & 17,354 & 0.01 & Harassment, Not Harassment  &  \citet{bretschneider2016detecting}, 6 \\\hline

Lexicon for Harassment & 2018 & \href{https://github.com/Mrezvan94/Harassment-Corpus} {GitHub} & Twitter & 24189 & 0.01 & Racism, Sexism, Appearance-related, Intellectual, Political  &  \citet{rezvan2018quality}, 19 \\\hline
 
Aggression and Friendliness & 2017 & \href{https://github.com/ewulczyn/wiki-detox} {GitHub} & Wikipedia & 160,000 & NA &  Very aggressive, Neutral, Very friendly  &  \citet{wulczyn2017ex}, 347\\\hline

   \end{tabular}}
\end{table}

\begin{table}
    \centering
    \caption{Hate speech datasets and Corpus  for Arabic, German,  Danish, French, Indonesian, Italian, Bengali, Urdu, Russian and Hindi.} 
    \label{tab:multilingualdatasets} 
    \scalebox{.75}{\begin{tabular}{|p{4cm}|p{.8cm}|p{.8cm}|p{1.5cm} |p{1cm}|l|p{5cm}|p{1.7cm}|p{3cm}|}
    \hline
    \bf{Name} &  \bf{Year} & \bf{Link} & \bf{Platform} & \bf{Size}  & \bf{Ratio} & \bf{Class} &\bf{Lang.} & \bf{Ref.} \\ \hline
    
Abusive Language Detection on Arabic Social Media & 2017 & \href{http://alt.qcri.org/~hmubarak/offensive/TweetClassification-Summary.xlsx} {Link} &  Twitter &   1,100 & 0.59 &  Obscene, Offensive but not obscene, Clean  & Arabic &\citet{mubarak2017abusive}, 128 \\\hline

Abusive Language Detection on Arabic Social Media & 2017 & \href{http://alt.qcri.org/~hmubarak/offensive/AJCommentsClassification-CF.xlsx} {Link} &  AlJazeera &    32,000 & 0.81 &  Obscene, Offensive but not obscene, Clean  & Arabic &  \citet{mubarak2017abusive}, 128 \\\hline

 Religious Hate Speech in the Arabic & 2018 & \href{https://github.com/nuhaalbadi/Arabic\_hatespeech}  {GitHub} &  Twitter & 16,914 & 0.45 & Hate, Not hate & Arabic  & \citet{Albadi2018are}, 53 \\ \hline
 
 Anti-Social Behaviour in Online Communication  & 2018 & \href{https://onedrive.live.com/?authkey=!ACDXj_ZNcZPqzy0id=6EF6951FBF8217F9!105cid=6EF6951FBF8217F9} {Link}  &  YouTube &   15,050 & .39 &  Offensive, Not Offensive& Arabic  &  \citet{alakrot2018dataset}, 26 \\\hline
 
 Multi-Aspect Hate Speech Analysis  & 2019 & \href{https://github.com/HKUST-KnowComp/MLMA\_hate\_speech}  {GitHub} &  Twitter &  3,353 & 0.64 &  Gender, Sexual orientation, Religion, Disability & Arabic &  \citet{ousidhoum2019multilingual}, 41 \\\hline
 
 Arabic Levantine HateSpeech Dataset & 2019 & \href{https://github.com/Hala-Mulki/L-HSAB-First-Arabic-Levantine-HateSpeech-Dataset} {GitHub} &  Twitter &   5,846 & .38 &  Hate, Abusive, Normal  & Arabic &  \citet{mulki2019hsab}, 34 \\\hline
    
European Refugee Crisis & 2017 & \href{https://github.com/UCSM-DUE/IWG_hatespeech_public} {GitHub} & League of Legends & 469 & NA & Anti-refugee hate, Not Hate & German &  \citet{ross2017measuring}, 219 \\\hline

Offensive Statements Towards Foreigners & 2016 & \href{http://ub-web.de/research/} {GitHub} &  Facebook & 5,836 & 0.11 & slightly offensive, explicitly offensive.  targets (Foreigner, Government, Press, Community, Other, Unknown) & German &  \citet{bretschneider2017detecting}, 21 \\\hline

GermEval 2018 & 2016 & \href{https://github.com/uds-lsv/GermEval-2018-Data} {GitHub} & Twitter & 8,541 & 0.34 & Offense, Other, Abuse, Insult, Profanity & German &  \citet{wiegand2018overview}, 152 \\\hline

HASOC-2019 (German) & 2019& \href{https://hasocfire.github.io/hasoc/2019/dataset.html} {GitHub} & Twitter, Facebook & 4,669 & 0.24 & Hate, Offensive,  neither, Hate, Offensive, or Profane & German &  \citet{mandl2019overview}, 120 \\\hline

Offensive Language and Hate Speech Detection for Danish & 2019 & \href{https://figshare.com/articles/Danish_Hate_Speech_Abusive_Language_data/12220805} {GitHub} &  Twitter, Reddit, newspaper comments &  3,600 & .12 & Offensive, Not, Within Offensive (Target, Not), Within Target (Individual, Group, Other)
 & Danish &  \citet{sigurbergsson2019offensive}, 54 \\\hline
 
CONAN HS French & 2019 & \href{https://github.com/marcoguerini/CONAN} {GitHub} & Synthetic, Facebook & 17,119 & 1 & Islamophobic,  not Islamophobic & French &  \citet{chung2019conan}, 31 \\\hline

MLMA hate speech & 2019 & \href{https://github.com/HKUST-KnowComp/MLMA_hate_speech} {GitHub} & League of Legends &  4,014 & 0.72 & Gender, Sexual orientation, Religion, Disability & French &  \citet{ousidhoum2019multilingual}, 41 \\\hline
Offensive Language Identification in Greek & 2020 & \href{https://sites.google.com/site/offensevalsharedtask/home} {GitHub} & Twitter &  4779 & 0.01 & Offensive, Not, Target, Not, Individual, Group, Other & Greek & \citet{pitenis2020offensive}, 41 \\\hline

HS in  Indonesian Language & 2017 & \href{https://github.com/ialfina/id-hatespeech-detection} {GitHub} & Twitter & 713 & 0.36 & Hate, Not Hate & Indonesian &  \citet{alfina2017hate}, 62 \\\hline

HS and Abusive Language in Indonesian Twitter & 2019 & \href{https://github.com/okkyibrohim/id-multi-label-hate-speech-and-abusive-language-detection} {GitHub} & Twitter & 13,169 & 0.42 & No hate speech, No hate speech but abusive, Hate speech but no abuse, Hate speech and abuse, Religion/creed, Race/ethnicity, Physical/disability, Gender/sexual orientation, Other invective/slander, within hate, strength (Weak, Moderate and Strongt) & Indonesian &  \citet{ibrohim2019multi}, 34 \\\hline

Preliminaries Study for Abusive Language & 2016 & \href{https://github.com/okkyibrohim/id-abusive-language-detection} {GitHub} & Twitter & 2,016 & 0.54 & Not abusive, Abusive but not offensive, Offensive & Indonesian &  \citet{ibrohim2018dataset}, 37 \\\hline

An Italian Twitter Corpus & 2018 & \href{https://github.com/msang/hate-speech-corpus} {GitHub} & Twitter & 1,827 &0.13 & Immigrants, Not & Italian &  \citet{sanguinetti2018italian}, 77 \\\hline

Hindi-English Code-mixed Data & 2018 & \href{https://github.com/kraiyani/Facebook-Post-Aggression-Identification} {GitHub} & Facebook & 21,000 & 0.27 & None, Covert Aggression, Overt Aggression, Physical threat, Sexual threat, Identity threat, Non-threatening aggression, Attack, Defend, Abet & Hindi, English &  \citet{kumar2018aggression}, 75 \\\hline

Hindi-English Code-mixed Data & 2018 & \href{https://github.com/kraiyani/Facebook-Post-Aggression-Identification} {GitHub} & Facebook & 18,000 & 0.06 & None, Covert Aggression, Overt Aggression, Physical threat, Sexual threat, Identity threat, Non-threatening aggression, Attack, Defend, Abet & Hindi, English &  \citet{kumar2018aggression}, 75 \\\hline

Offensive Tweets in Hinglish Language& 2018 & \href{https://github.com/pmathur5k10/Hinglish-Offensive-Text-Classification} {GitHub} & Twitter & 3,189 & 0.65 & Not Offensive, Abusive, Hate & Hindi, English &  \citet{mathur2018did}, 34 \\\hline

A Dataset of Hindi-English Code-Mixed & 2018 & \href{https://github.com/deepanshu1995/HateSpeech-Hindi-English-Code-Mixed-Social-Media-Text} {GitHub} & Twitter &  4,575 & 0.36 & Hate, Not & Hindi, English &  \citet{bohra2018dataset}, 56 \\\hline

HASOC-2019 (Hindi) & 2016 & \href{https://hasocfire.github.io/hasoc/2019/dataset.html} {GitHub} & Twitter, Facebook & 5,983 & 0.51 & Hate, Offensive or Neither,  Profane,  Targeted or Untargeted & Hindi, English &  \citet{mandl2019overview}, 120 \\\hline

Bengali HaS Dataset& 2020 & \href{https://github.com/rezacsedu/Bengali-Hate-Speech-Dataset} {GitHub} & Facebook, YouTube, Wikipedia, news-articles & - & - & Personal, Political, Religious, Geopoitical and 	Gender abusive hate & Bengali &  \citet{karim2020classification}, 4 \\\hline

Russian Dataset & 2020 & \href{https://github.com/bohdan1/AbusiveLanguageDataset} {GitHub} & - & - & - & - & Russian and Ukrainian  &  \citet{andrusyak2018detection}, 4 \\\hline

Roman Urdu & 2020 & \href{https://github.com/haroonshakeel/roman_urdu_hate_speech} {GitHub} & - & - & - & - & Urdu  &  \citet{u1}, 3 \\\hline

   \end{tabular}}
\end{table}

\clearpage

\begin{figure}
    \centering
    \includegraphics[width=.9\linewidth,]{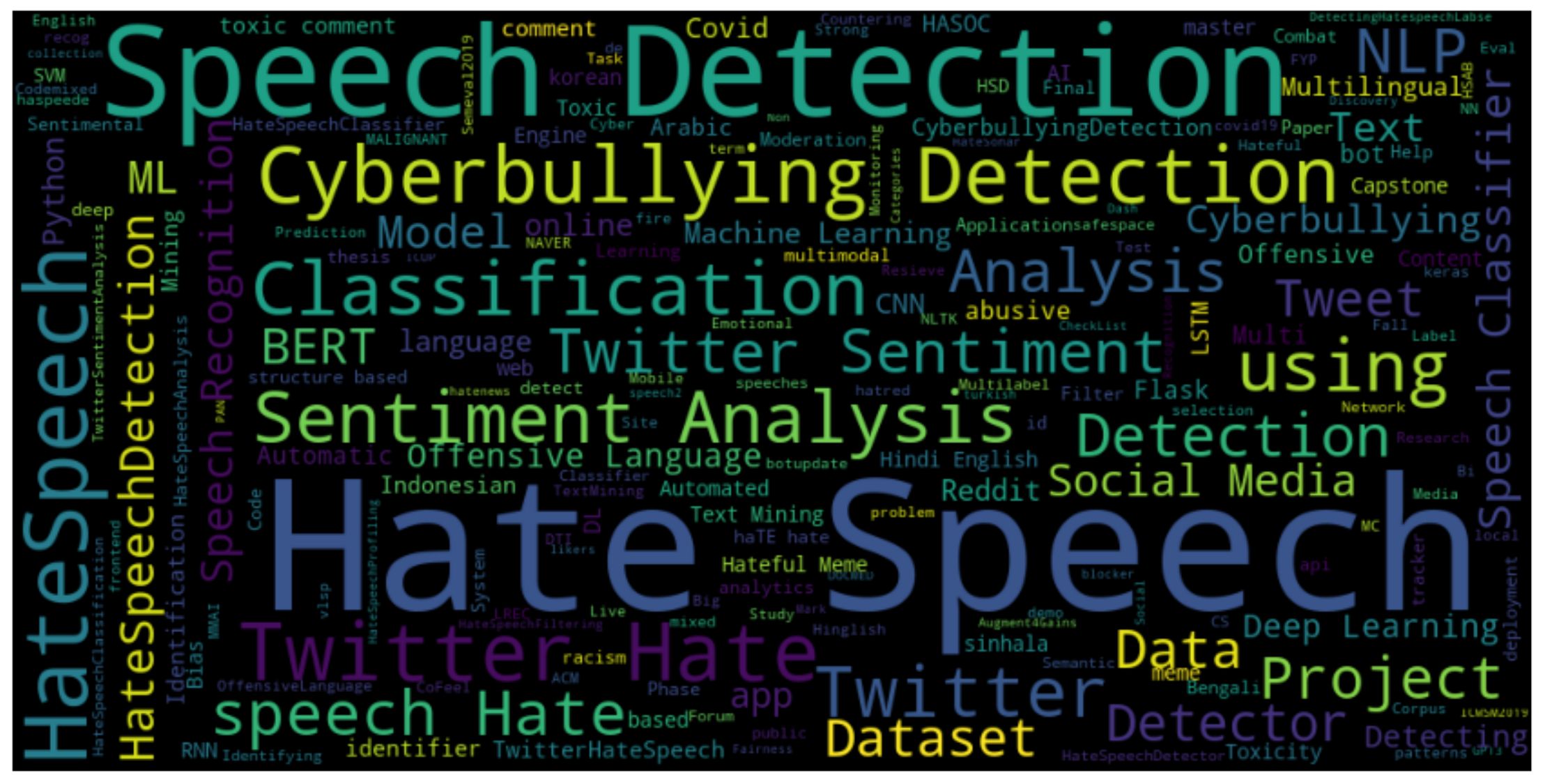}
    \caption{Word cloud representation of GitHub open source projects.}
    \label{fig:github_open}
\end{figure}

\begin{figure}
    \centering
    \includegraphics[width=.6\linewidth,]{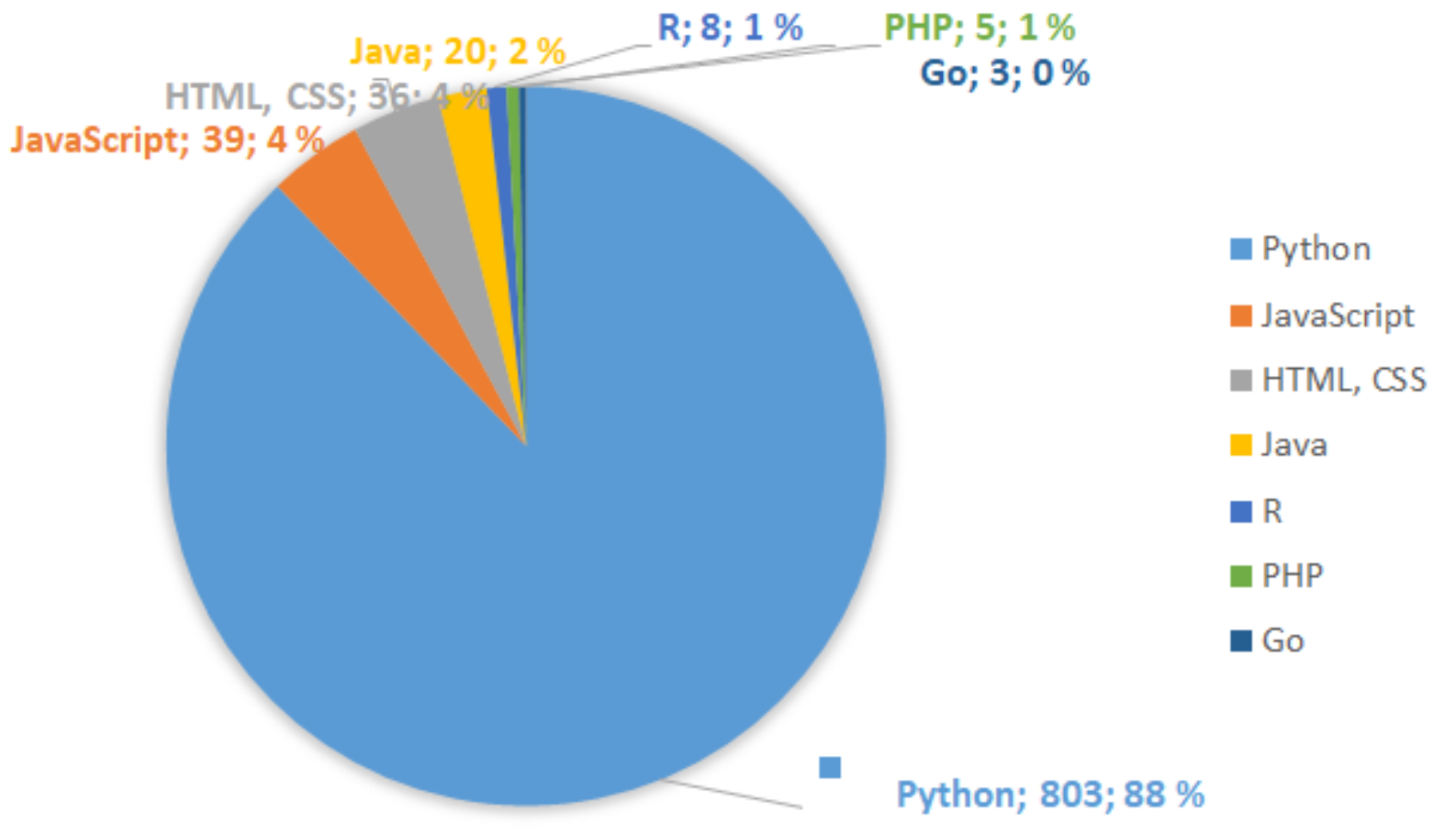}
    \caption{GitHub open source projects programming language.}
    \label{fig:programming}
\end{figure}

\subsection{Open source projects}
\label{open_project}
We checked if there are any open-source projects available for hate-speech automatic detection or can be used as examples or sources for annotated data. For this, we carried out a search on GitHub repository with the search query "hate speech" in the available search engine. We found 1039 repositories, and only 53 were regularly forked and updated. Since this is a large number of repositories, it was challenging to include all of them in this paper and comment on them individually. Therefore, we have restricted to the 15 top-ranked one. Furthermore, we have exported the project repository names and descriptions into a CSV file for word cloud representation, which may help us understand the content of these open source projects in terms of the provided description.

Table~\ref{tab:opensource} shows some highly cited HS detection papers source code. For example, \citeauthor{davidson2017automated} used Twitter dataset with TF-IDF, n-gram feature and LR-SVC model architecture. Furthermore, we have found the source code of \citeauthor{badjatiya2017deep} which used FastText and CNN, and LSTM models, achieving 78\% F1 score and 85\% accuracy.  Furthermore, a new Korean dataset found in  highly 'forked' GitHub repository claimed to be the first human-annotated Korean corpus for toxic speech detection and sizeable unlabeled corpus (Tab. \ref{tab:opensource}, Index 2). 

Another interesting repository named "Hate\_sonar" used the BERT approach and the dataset in \citeauthor{davidson2017automated}. It created an easily installable python library, which anyone can use for their test project without having any coding skill.

Furthermore,  some highly 'started' and 'forked'  works appeared mainly relevant to sentiment analysis; namely TextBlob, VaderSentiment and Transformer. Here, the Transformer provides thousands of pre-trained models (mainly BERT) to perform tasks on texts such as classification, information extraction, question answering, summarization, translation, text generation, and sentiment analysis \cite{wolf-etal-2020-transformers}.

Figure~\ref{fig:github_open} shows the word cloud representation of the identified GitHub project descriptions. The illustration indicates the high proportion of words such as hate speech, hateSpeech, speech hate, hateful meme, cyberbullying offensive language, and toxic comment in the repositories related to HS detection. Furthermore, Twitter, tweet, Twitter hate, Reddit, and other social media were also present in this illustration, although with less intensity compared to hate detection related wording. Similarly, deep-learning methods such as BERT, LSTM, CNN, where the increased popularity of BERT is more emphasized. Surprisingly, this is in a full agreement with the literature review as well, where BERT was found to be a dominant trend in deep-learning methods. Among the fifteen top forked repositories, four projects were linked to BERT based model. The connection between hate speech and sentiment analysis is also stressed in Fig.~\ref{fig:github_open}, as well as the growing interest in non-English HS detection through wordings like multilingual, Indonesian, Hindi, indicating repositories for multilingual HS detection tasks.

Regarding the programming languages employed, we noticed that most of the projects (88\%) were developed in Python language, while others used JavaScript (4\%), Java (2\%), HTML and CSS (4\%), and three projects with GO (Fig.~\ref{fig:programming}).

\begin{table}[ht]
    \centering
    \caption{Fifteen most popular GitHub open source projects.} 
    \label{tab:opensource} 
    \scalebox{.75}{\begin{tabular}{|p{3cm}|p{2cm}|p{6.5cm}|p{2cm} |p{1.7cm}|p{1.7cm}|p{.7cm}|p{.7cm}|}
    \hline
    \bf{Repository Name}& \bf{Github Link} &  \bf{Focus} &\bf{Publication Ref., Year}& \bf{Features Representation} & \bf{Algorithm}  &  \bf{Star}& \bf{Fork}  \\ \hline
    
1. Hate speech and offensive language & \href{https://github.com/t-davidson/hate-speech-and-offensive-language}{Link}, Source code given & Twitter 25k dataset used for HS detection.  & \citet{hateoffensive}, \citeyear{hateoffensive} & TF-IDF, n-gram, bi-gram & LR, SVC & 545 & 217 \\\hline

2. Korean HateSpeech Dataset & \href{https://github.com/kocohub/korean-hate-speech}{Link}, Dataset source code given but project source code not available. & The first human-annotated Korean corpus for toxic speech detection and the large unlabeled corpus.
The data is comments from the Korean entertainment news aggregation platform. & \citet{moon-etal-2020-beep}, \citeyear{moon-etal-2020-beep} & -& - & 195 & 12\\\hline

3. Twitter hatespeech & \href{https://github.com/pinkeshbadjatiya/twitter-hatespeech}{Link}, source code available. & Implementation of  paper - "Deep Learning for Hate Speech Detection" & \citet{badjatiya2017deep}, \citeyear{badjatiya2017deep}& Fasttext, BOW & CNN, LSTM & 182 & 75\\\hline

4. Hate sonar & \href{https://github.com/Hironsan/HateSonar}{Link}, Install \$ pip install hatesonar & HateSonar allows you to detect hate speech and offensive language in text, without the need for training. There's no need to train the model. You have only to fed text into HateSonar. It detects hate speech with the confidence score.  & \citet{davidson2017automated}, \citeyear{davidson2017automated} & - & BERT & 119 & 23\\\hline

5. Hate speech dataset & \href{https://github.com/Vicomtech/hate-speech-dataset}{Link}, Source code available & These files contain text extracted from Stormfront, a white supremacist forum. A random set of forums posts have been sampled  and n manually labelled as containing hate speech or not. & \citet{gibert2018hate}, \citeyear{gibert2018hate} & - &   & 68 & 38\\\hline

6. Dataset for Learning to Intervene
 & \href{https://github.com/jing-qian/A-Benchmark-Dataset-for-Learning-to-Intervene-in-Online-Hate-Speech}{Link}, Only dataset available &  HS intervention along with two fully-labeled datasets collected from Gab and Reddit. Distinct from existing hate speech datasets, their datasets retain their conversational context and introduce human-written intervention responses. &  & - &   & 40 & 7\\\hline

7. HateXplain
 & \href{https://github.com/hate-alert/HateXplain}{Link},  dataset and soruce code available, Install: pip install -r requirements.txt  &  Multilingual multi-aspect hate
speech analysis dataset & \citet{mathew2020hatexplain}, \citeyear{mathew2020hatexplain} & - &  BERT & 39 & 8\\\hline

8. MLMA hate speech
 & \href{https://github.com/HKUST-KnowComp/MLMA_hate_speech/blob/master/predictors.py}{Link},  dataset and soruce code available  &  Multilingual multi-aspect hate
speech analysis dataset & \citet{ousidhoum-etal-multilingual-hate-speech-2019}, \citeyear{ousidhoum-etal-multilingual-hate-speech-2019}& - &  LR & 35 & 3\\\hline

9. Korean Hate Speech
 & \href{https://github.com/inmoonlight/detox}{Link},  dataset and soruce code available  &  This hate speech detection model trained on kocohub/korean-hate-speech & 2020 & - &  BERT & 34 & 4\\\hline

10. likers-blocker
 & \href{https://github.com/dmstern/likers-blocker}{Link},  Browser addon  &  Block hate promoter twitter suer & 2021 & - &   & 28 & 0\\\hline

11. Twitter Hate Speech Detection
 & \href{https://github.com/vedant-95/Twitter-Hate-Speech-Detection}{Link},  dataset and soruce code available  &  project analyzed a dataset CSV file from Kaggle containing 31,935 tweets. & 2020  & BOW, TF-IDF &  LR, NB, RF & 26 & 16 \\\hline

12. Korean hate speech language modeling
 & \href{https://github.com/captainnemo9292/hate-speech-language-modeling}{Link},  dataset and soruce code available  &  Recurrent Neural Network based Hate Speech Language Model for Korean Hate Speech Detection & 2020 & &  RNN & 26 & 16\\\hline
 
13. TextBlob
 & \href{https://github.com/sloria/TextBlob}{Link},  soruce code available, install: pip install -U textblob  &  TextBlob is a Python library provides a simple API for diving into common natural language processing (NLP) tasks such as part-of-speech tagging, noun phrase extraction, sentiment analysis, classification, translation, and more & 2021 & &  & 7.6k & 1k\\\hline
 
14. vaderSentiment
 & \href{https://github.com/cjhutto/vaderSentiment}{Link},  soruce code available, install: pip install vaderSentiment  & VADER  is a lexicon and rule-based sentiment analysis tool that is specifically attuned to sentiments expressed in social media. & 2021 & &  & 2.9k & 767\\\hline
 
 15. Transformers
 & \href{https://github.com/huggingface/transformers}{Link},  soruce code available, install: pip install transformers  & Transformers provides thousands of pre-trained models to perform tasks on texts such as classification, information extraction, question answering, summarization, translation, text generation, sentiment analysis & \citet{wolf-etal-2020-transformers}, \citeyear{wolf-etal-2020-transformers} & & BERT, ALBERT & 29k & 8.7k\\\hline

\end{tabular}}
\end{table}

\clearpage

\section{Research challenges and opportunities}
\label{challenges}

The above literature review for deep learning and non-deep learning and resource analysis summarized the main research in the field of HS automatic detection from textual inputs. At the same time, we have also identified several challenges and research gaps (Table~\ref{tab:reviewGap}) from previous research.
 
\subsection{Open Source Platforms or Algorithms:} There are indeed many open-source projects available related to HS. However, only few project source codes are available from well-known publications. From the 1039 projects in GitHub, we have only found 53 projects regularly maintained and forked, which may question the usability and source code quality of the rest of the projects. More sharing of code with a clear documentation, algorithms, processes for feature extraction, and open-source datasets can help the discipline evolves more quickly.

\subsection{Language and System Barriers}
Language evolves quickly,  particularly among young populations that frequently communicate in social networks, demanding continuity of research for HS datasets.  For instance, online platforms are removing hate contents manually and automatically \footnote{\url{https://about.fb.com/news/2021/02/update-on-our-progress-on-ai-and-hate-speech-detection/}} \footnote{\url{https://blog.twitter.com/en_us/topics/company/2019/hatefulconductupdate.html}}. However, those who spread HS content will always try to develop a new way to evade and by pass any system imposed restriction. For example, some users do post HS content as images containing the hate text, which circumvent some basis automatic HS detection. Although image to text conversion might solve some particular problem, still several challenges arise due to limitation of such conversation as well as existing automatic HS detection. Besides, changing the language structure could be another challenge, for example,  through usage of unknown abbreviations and mixing different languages, e.g., i) Writing part of a sentence in one language and the other part in another language; (ii) Writing sentence phonetics in another language (e. g., writing Hindi sentences using English).

\subsection{Dataset:}. There are no commonly accepted datasets recognized as ideal for automatic HS detection task. Authors annotate dataset differently based on their understanding and tasks requirement. For instance, Figure \ref{fig:class} highlights  47 different annotation labels from 69 datasets, which stress on the diversity of the existing datasets.  Besides, 55\% of the available datasets' sizes are small and contain a tiny portion of hate content. Many datasets were annotated through crowd-sourcing, which may also question the knowledge of the annotator. 

\textbf{ Clear label definitions}. There is a prerequisite to have a clear label definition, separating HS from other types of offensive languages \cite{davidson2017automated} \cite{founta2018large}. Indeed, dataset can cover a broader spectrum targeting multiple fine-grained HS categories (e.g., sexism, racism, personal attacks, trolling, cyberbullying). This can be performed through either multi-labelling approach, although one notices the presence of ambiguous cases as in \citeauthor{waseem2016you}’s racism and sexism labels, or in a hierarchical manner as in \citeauthor{basile2019semeval}'s and \citeauthor{kumar2018aggression}'s work on subtypes of HS and aggression, respectively. 

\textbf{Annotation quality}. The offensive nature of hate speech and abusive language makes the grammatical structure and cross-sentence boundaries loose, leading to challenging annotation criteria \cite{nobata2016abusive}. Therefore, hate speech datasets should be constantly updated according to newly available knowledge. For instance, \citeauthor{Poletto} found that only about two-thirds of the existing datasets report inter-annotator agreement, guidelines, definitions, and examples. To ensure a high inter-annotator agreement, extensive instructions and the use of expert annotators are required.
Furthermore, 98\% of the datasets were collected from social networks and labeled manually. Only limited work was directed towards (artificially) dataset creation and enrichment of existing datasets.

\subsection{Comparative Studies}
From Fig.~\ref{fig:statistic_all_algorithm}, we identified 24 different hybridization schemes in deep-learning models. However, extensive and comprehensive comparative studies where different approaches are genuinely contrasted and compared were very missing. This leave the door widely open to future comprehensive comparative HS studies in terms of data preprocessing, feature engineering, model training and evaluation.  

In addition, little to none work has been found focusing on the labelling issue taking into account the model organization of the individual/group targeted by the HS post, building from advances in social psychology theory and human computer interaction research. This creates a gap in the practicality of the automatic detection HS developed models as well as on the comparison analysis. This also raises many questions about how the underlined HS detection technique would impact real users' experience and the accuracy of the model compared to human moderation. To answer this question, more interdisciplinary studies and collaboration with organizations are needed. 

Besides, comparison between various training approaches, debiasing approaches, overfitting models, and what characteristics of the datasets interact with the effectiveness would be worth investigating. For example, when performing transfer learning, the trade-off between domain-specificity, linguistic patterns, and underlying sentiment of hate speech can be considered before model design, feature extraction and preprocessing.

\subsection{Multilingual Research}  As previously pointed out, 50\% of HS studies, datasets, and open source projects were provided in English language. Although, we noticed arise in some other non-English resources as well, as in Arabic where AraVec word embedding showed some popularity, there is a scarcity in the development of other non-English NLP resources. Although, we have identified 21 different language HS related works, which creates an opportunity to develop enhanced NLP tools for these languages.

\begin{table}[ht]
    \centering
    \caption{Review of Gaps and future research agenda.} 
    \label{tab:reviewGap} 
    \scalebox{.90}{\begin{tabular}{|p{4cm}|p{4cm}|p{9cm}|}
    \hline
    \bf{Research question} &  \bf{Gap in literature} &  \bf{Future Research Agenda}  \\ \hline
 Q1: What are the specificities among different HS branches and scopes for automatic HS detection from previous literature? 
 
 & G1: Discrepancy and fragmentation of knowledge among different domain \newline  \newline \newline \newline \newline \newline 
 G2: Multilingual resources. 
 
 & - How to develop a common framework for researchers which will be helpful for domain adaptation? \newline 
 - How we can clarify all the concepts and definitions that will be helpful to obtain high quality and comparable resources? \newline 
 - How effectively take into account the specificities related to language and culture, and work towards preventing HS? \newline \newline - Development of  NLP resources for other languages (e.g., multilingual sentiment feature, dataset, word embedding, etc.) would leverage HS detection for other languages. \\\hline
 
  Q2: What is the state of deep learning in automatic HS detection in practice? 
  
  & G3: Comparative study among different deep learning algorithms and related resources. \newline\newline\newline\newline\newline  
   G4: Model application and impact 
  
  & - Which existing deep learning models, features, and what characteristics of the datasets  are more efficient in tackling HS detection?  \newline 
  - What approaches could be used to make the model less biased against specific terms or language styles, from the perspectives of training data or objective.  More systematic comparisons between debiasing approaches would be favorable. \newline\newline  
  - Use of deep-learning architecture to create NLP resources which are currently developed with non-deep learning methods that will leverage HS detection. \newline  
  - Can automatic deep-learning models practically aid human moderators in content moderation? In that case, how can human moderators or organizations make use of the outputs feature analysis most effectively? Would that introduce more bias or reduce bias in the content moderation process? What would the impact be on the users of the platform? To answer these questions, interdisciplinary study and collaboration with organizations are needed. \\\hline
  
  Q3: What is the state of the HS datasets in practice? 
  
  & G5: Dataset annotation   
  \newline\newline\newline\newline\newline\newline\newline \newline  \newline\newline\newline\newline
  G6: Dataset augmentation tools and techniques 
  
  & - How to avoid the risk of creating data that are biased or too much related to a specific resource? 
  \newline - Development of data-driven taxonomy that highlights how different types of HS datasets concepts are linked and how they differ from one another? 
  \newline - How to deal with the discrepancy of data and error analysis of human annotation issued for previous literature? 
  \newline - How to develop a standard form of annotation guideline? \newline - What type of and how much training or instruction is required to match the annotations of crowdworkers and experts?
  \newline \newline - Develop new methods or algorithms for artificial HS dataset creation and expand the semantic meaning of existing datasets that will help to create a large-scale balanced dataset. \\\hline
   \end{tabular}}
\end{table}
\clearpage

\section{Conclusion}
\label{Conclusion}
In this survey, we presented a critical overview of how the automatic detection of hate speech in the text has evolved over the past few years. Our analysis also included other hate speech domains, e.g., cyberbullying, abusive language, discrimination, sexism, extremism and radicalization.
We initially reviewed existing surveys in the field. We found few recent systematic literature reviews related to HS detection, which are shown to be insufficient to summarize the current state of research in the field.
Next, we carried out a systematic literature review from Google scholar and ACM digital library databases for all documents related to hate speech published between 2000 and 2021. A total of 463 articles are found to match PRISMA inclusion and exclusion criteria. The findings indicate that initially SVM algorithm and various types of TF-IDF features were the most widely used. However, after the advancement in deep-learning technology, a rapid change in the hate speech analysis methods was observed. The research community preferred to use different kinds of word embedding with CNN and RNN architectures. From 2017 to 2021, several comparative studies have shown the merits of deep-learning models including CNN, RNN using word2Vec, GloVe, FastText, among other embedding as compared to traditional machine learning models such as SVM, LR, NB, and RF models. Nevertheless, comparison among deep learning models is still in its infancy. For instance, mixed results were obtained when comparing CNN to RNN models. Though some authors opined RNNs are more suited for the long-ranged context, while CNN seem useful in extracting features and GRU is found to be more suited for long sentences. Besides, multiple papers suggested that the concatenation of two or more deep learning models performed better than using a single deep learning model. For example, CNN+LSTM and CNN+GRU both performed better than the single application of LSTM and CNN.

Similarly, the comparison of different word-embedding models, e.g, FastText, Word2Vec, GloVe, showed a close performance; however, ELMO performed slightly better than others. Though the work related to ELMO is meager, which leaves the door open for further comparative studies.  After introducing contextual relations-based model BERT in 2018,  several works claimed BERT's outperforming ELMO, CNN and RNN models. After comparing the best HS detection architecture from Semeval-2019, Semeval-2020 and the HASOC-2020 competitions, BERT-based model was also ranked top among other deep-learning models. 

In the final part, we analyzed 69 hate speech datasets. The existing works presented several obstacles for dataset preparation. In general, researchers tend to start by collecting and annotating new comments from SM or using previous datasets. Often, retrieving an old dataset from Twitter is not always fully possible due to tweets' potential removal. This slows down the research's progress because less data is available, making it more challenging to compare different studies' results.  Several limitations are found when scrutinizing statistics of past dataset. Significantly, most of the dataset sizes were small, lack the ratio of hate content, and lack label definitions and inter-annotator agreements. Besides, only a fraction of two datasets were synthetically made, which leaves room for artificially dataset creation, augmentation, and enrichment.
Finally, we identified the main challenges and opportunities in this field. This includes the scarcity of good open-source code that is regularly maintained and used by the society, the lack of comparative studies that evaluate the existing approaches, and the absence of resources in non-English experiments. With our work, we summarized the current state of the automatic HS detection field. Undoubtedly, this is an area of profound societal impact and with many research challenges.

\section*{Acknowledgments}
This work is supported by the European Young-sters Resilience through Serious Games, under the Internal Security Fund-Police action:823701-ISFP-2017-AG-RAD grant, which is gratefully acknowledged.

\bibliographystyle{cas-model2-names}

\bibliography{references}


\end{document}